%% file: main.tex
\documentclass{article} 
\usepackage{iclr2022_conference,times}

\usepackage[OT1]{fontenc} 

\input{math_commands.tex}

\usepackage{hyperref}
\usepackage{url}
\usepackage{graphicx}
\usepackage{amsmath}
\usepackage{amssymb}
\usepackage{subcaption}
\usepackage{amsfonts}
\usepackage{booktabs}
\usepackage{siunitx}
\usepackage{pifont}

\title{Robust Models Are More Interpretable \\ Because Attributions Look Normal}

\iclrfinalcopy

\author{Zifan Wang, Matt Fredrikson \& Anupam Datta \\
Carnegie Mellon Univeristy\\
Pittsburgh, PA 15213, USA \\
\texttt{zifan@cmu.edu} \\
}

\begin{document}

\input{macros}
\input{defs}
\maketitle

\input{00_Abstract}


\input{01_Introduction}

\input{02_Background}
\input{03_Method}

\input{04_Evaluation}

\input{05_Application}
\input{06_RelatedWork}

\input{07_Conlusion}



\section*{Acknowledgement}

This work was developed with the support of NSF grant CNS-1704845 as well as by DARPA and the Air Force Research Laboratory under agreement number FA8750-15-2-0277. The U.S. Government is authorized to reproduce and distribute reprints for Governmental purposes not with- standing any copyright notation thereon. The views, opinions, and/or findings expressed are those of the author(s) and should not be interpreted as representing the official views or policies of DARPA, the Air Force Research Lab- oratory, the National Science Foundation, or the U.S. Government.

\bibliography{iclr2022_conference}
\bibliographystyle{iclr2022_conference}

\appendix
\input{08_Appendix.tex}

\end{document}

%% file: math_commands.tex
\usepackage{amsmath,amsfonts,bm}

\def\eqref#1{equation~\ref{#1}}

\def\1{\bm{1}}

\DeclareMathAlphabet{\mathsfit}{\encodingdefault}{\sfdefault}{m}{sl}
\SetMathAlphabet{\mathsfit}{bold}{\encodingdefault}{\sfdefault}{bx}{n}

\newcommand{\E}{\mathbb{E}}



%% file: macros.tex
\newcommand{\outline}[1]{\textcolor{blue}{#1}}

\newcommand{\xvec}{\mathbf{x}}
\newcommand{\zvec}{\mathbf{z}}
\newcommand{\wvec}{\mathbf{w}}
\newcommand{\bvec}{\mathbf{b}}
\newcommand{\nvec}{\mathbf{n}}

%% file: defs.tex

\newcommand{\madry}[0]{M\k{a}dry}

\newcommand{\premath}[0]{\vspace{-0.1in}}

\newcommand{\const}[1]{\text{#1}}
\newcommand{\dataset}[0]{\mathcal{D}}
\newcommand{\indicator}[1]{\mathds{1}\paren{#1}}

\newcommand{\inaive}[0]{g_\text{naive}}
\newcommand{\igrammar}[0]{g_\text{grammar}}
\newcommand{\wdebias}[0]{$\text{WED}$\xspace}
\newcommand{\wdebiasbefore}[0]{$\overleftarrow{\wdebias}$\xspace}
\newcommand{\wdebiasafter}[0]{$\overrightarrow{\wdebias}$\xspace}
\newcommand{\obias}[0]{$\text{OB}$}
\newcommand{\aobias}[0]{$\text{AOB}$}

\newcommand{\norm}[1]{\left\lVert#1\right\rVert}



\newtheorem{theorem}{Theorem}
\newtheorem{definition}{Definition}
\newtheorem{lemma}{Lemma}
\newtheorem{example}{Example}
\newtheorem{property}{Property}
\newtheorem{corollary}{Corollary}
\newtheorem{proposition}{Proposition}
\newtheorem{claim}{Claim}
\newtheorem{remark}{Remark}
\newtheorem{axiom}{Axiom}


\newcommand{\stacklabel}[1]{\stackrel{\smash{\scriptscriptstyle \mathrm{#1}}}}
\newcommand{\defeq}{\stacklabel{def}=}
\newcommand{\defeqq}{\stacklabel{def?}=}
\newcommand{\tab}{\hspace{0.5in}}



\newcommand{\abs}[1]{\left| #1 \right|}
\newcommand{\paren}[1]{\left( #1 \right)}
\newcommand{\sparen}[1]{\left[ #1 \right]}
\newcommand{\vparen}[1]{\left< #1 \right>}
\newcommand{\set}[1]{\left\{ #1 \right\}}
\newcommand{\qm}[1]{``#1''}
\newcommand{\vect}[1]{\left\langle #1 \right\rangle}

\newcommand{\nexist}{\cancel{\exists}}
\newcommand{\nin}{{\; \cancel \in \;}}

\newcommand{\Ra}{\Rightarrow}
\newcommand{\la}{\leftarrow}
\newcommand{\La}{\Leftarrow}

\newcommand{\real}{\mathbb{R}}
\newcommand{\integer}{\mathbb{Z}}
\newcommand{\setsize}[1]{\left|#1\right|}


 \newcommand{\prob}[2]{\Pr_{#1}[#2]}
\newcommand{\expect}[1]{\E\paren{#1}}
\newcommand{\expectsub}[2]{\E_{#1}\sparen{#2}}
\newcommand{\given}[0]{\; | \;}






%% file: 00_Abstract.tex

\begin{abstract}
    Recent work has found that adversarially-robust deep networks used for image classification are more interpretable: their feature attributions tend to be sharper, and are more concentrated on the objects associated with the image's ground-truth class.
    We show that smooth decision boundaries play an important role in this enhanced interpretability, as the model's input gradients around data points will more closely align with boundaries' normal vectors when they are smooth.
    Thus, because robust models have smoother boundaries, the results of gradient-based attribution methods, like Integrated Gradients and DeepLift, will capture more accurate information about nearby decision boundaries.
    This understanding of robust interpretability leads to our second contribution: \emph{boundary attributions}, which aggregate information about the normal vectors of local decision boundaries to explain a classification outcome.
    We show that by leveraging the key factors underpinning robust interpretability, boundary attributions produce sharper, more concentrated visual explanations---even on non-robust models. Any example implementation can be found at \url{https://github.com/zifanw/boundary}.

    
 \end{abstract}

%% file: 01_Introduction.tex

\section{Introduction}\label{sec:introduction}

\emph{Feature attribution methods} are widely used to explain the predictions of neural networks~\citep{binder2016layer, dhamdhere2018how, 8237633, leino2018influence, montavon2015explaining, selvaraju2017grad, shrikumar2017learning, simonyan2013deep,smilkov2017smoothgrad,springenberg2014striving,sundararajan2017axiomatic}.
By assigning an importance score to each input feature of the model, these techniques help to focus attention on parts of the data most responsible for the model's observed behavior.
Recent work~\citep{croce2019provable, pmlr-v97-etmann19a} has observed that feature attributions in adversarially-robust image models, when visualized, tend to be more interpretable---the attributions correspond more clearly to the discriminative portions of the input.

One way to explain the observation relies on the fact that robust models do not make use of \emph{non-robust features}~\citep{NEURIPS2019_e2c420d9} whose statistical meaning can change with small, imperceptible changes in the source data.
Thus, by using only robust features to predict, these models naturally tend to line up with visibly-relevant portions of the image.
\citeauthor{pmlr-v97-etmann19a} take a different approach, showing that the gradients of robust models' outputs more closely align with their inputs, which explains why attributions on image models are more visually interpretable.

\begin{figure*}[!t]
    \centering
    \begin{subfigure}[b]{0.45\textwidth}
        \centering
        \includegraphics[width=0.95\textwidth]{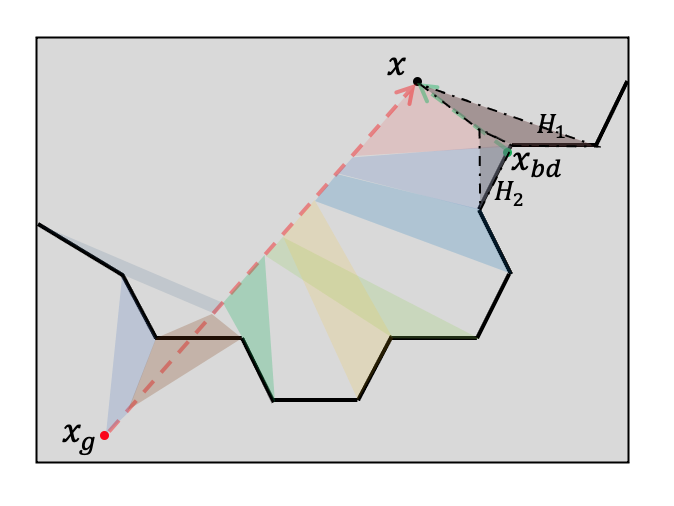}
        \caption{}
        \label{fig:exmaple-bd}
    \end{subfigure}
    \hfill
    \begin{subfigure}[b]{0.45\textwidth}
        \centering
        \includegraphics[width=1.1\textwidth]{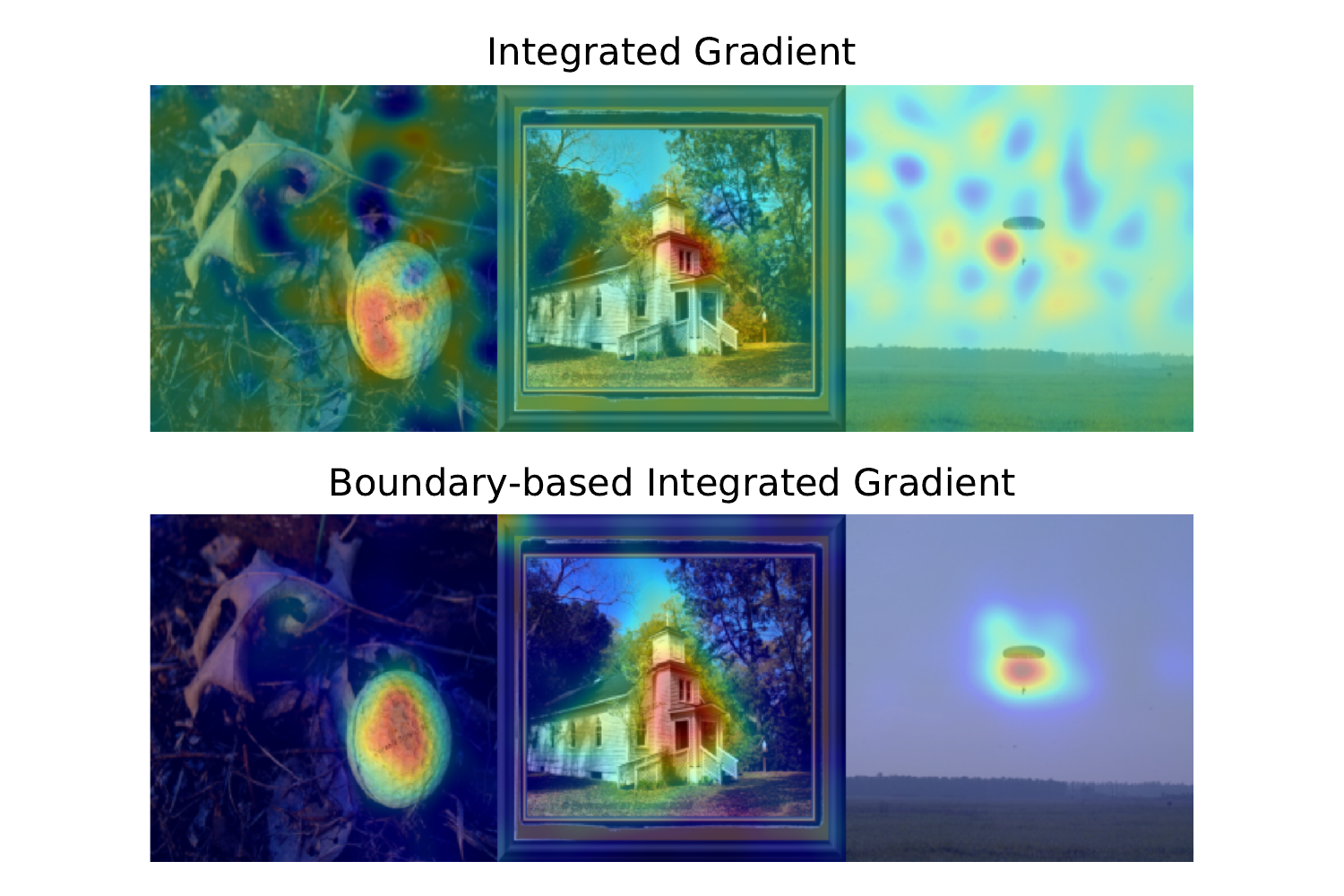}
        \caption{}
        \label{fig:exmaple}
    \end{subfigure}
       \caption{(a) Visualizations of geometrical interpretations of Saliency Map (SM), Boundary-based Saliency Map (BSM), Integrated Gradient (IG) and Boundary-based Integrated Gradient (BIG). Gradient computations can be viewed as projecting the input onto a particular decision boundary. While SM projects to a nearby boundary ($H_1$), BSM projects to the nearest one ($H_2$). IG (the red dashed path) from a global baseline $\xvec_g$, i.e. zeros, aggregates boundaries in colorful shaded areas; BIG (the green dashed path) integrates from the point $x_{bd}$ on the nearest boundary $H_2$ to $x$ and therefore aggregates nearby boundaries, $H_1$ and $H_2$ in gray shaded areas. (b) Visualizations of Integrated Gradient and the proposed improvement of it, Boundary-based Integrated Gradient, which is sharper, more concentrated and less noisy.}
\end{figure*}

In this paper, we build on this geometric understanding of robust interpretability.
With both analytical (Sec.~\ref{sec:method}) and empirical (Sec.~\ref{sec:evaluation}) results, we show that the gradient of the model with respect to its input, which is the basic building block of all gradient-based attribution methods, tends to be more closely aligned with the normal vector of a nearby decision boundary in robust models than in ``normal'' models. 
Leveraging this understanding, we propose Boundary-based Saliency Map (BSM) and Boundary-based Integrated Gradient (BIG), two variants of \emph{boundary attributions} (Sec.~\ref{sec:boundary-attributions}), which base attributions on information about nearby decision boundaries (see an illustration in Fig.~\ref{fig:exmaple-bd}). While BSM provides theoretical guarantees in the closed-form, BIG generates both quantitatively and qualitatively better explanations.
We show that these methods satisfy several desireable formal properties, and that even on non-robust models, the resulting attributions are more focused (Fig.~\ref{fig:exmaple}) and less sensitive to the ``baseline'' parameters required by some attribution methods.

To summarize, our main contributions are as follows. \emph{(1)} We present an analysis that sheds light on the previously-observed phenomeon of robust interpretability, showing that alignment between the normal vectors of decision boundaries and models' gradients is a key ingredient (Proposition~\ref{theorem-randomized-smoothing}, Theorem~\ref{theorem-robust-attributions}).
\emph{(2)} Motivated by our analysis, we introduce \emph{boundary attributions}, which leverage the connection between boundary normal vectors and gradients to yield explanations for non-robust models that carry over many of the favorable properties that have been observed of explanations on robust models.
\emph{(3)} We empirically demonstrate that one such type of boundary attribution, called \emph{Boundary-based Integrated Gradients} (BIG), produces explanations that are more accurate than prior attribution methods (relative to ground-truth bounding box information), while mitigating the problem of \emph{baseline sensitivity} that is known to impact applications of Integrated Gradients~\cite{sundararajan2017axiomatic} (Section~\ref{sec:application}).

%% file: 02_Background.tex

\section{Background}\label{sec:background}

We begin by introducing our notations. Throughout the paper we use italicized symbols $x$ to denote scalar quantities and bold-face $\xvec$ to denote vectors. 
We consider neural networks with ReLU as activations prior to the top layer, and a softmax activation at the top. 
The predicted label for a given input $\xvec$ is given by $F(\xvec) = \arg\max_c f_c(\xvec), \xvec \in \mathbb{R}^{d}$, where $F(\xvec)$ is the predicted label and $f_i(\xvec)$ is the output on the class $i$. 
As the softmax layer does not change the ranking of neurons in the top layer, we will assume that $f_i(\xvec)$ denotes the pre-softmax score. 
Unless otherwise noted, we use $||\xvec||$ to denote the $\ell_2$ norm of $\xvec$, and the $\ell_2$ neighborhood centered at $\xvec$ with radius $\epsilon$ as $B(\xvec, \epsilon)$. 

\paragraph{Explainability.} Feature attribution methods are widely-used to explain the predictions made by DNNs, by assigning importance scores for the network's output to each input feature.
Conventionally, scores with greater magnitude indicate that the corresponding feature was more relevant to the predicted outcome. 
We denote feature attributions by $\zvec = g(\xvec, f), \zvec, \xvec \in \mathbb{R}^{d}$. 
When $f$ is clear from the context, we simply write $g(\xvec)$. 
While there is an extensive and growing literature on attribution methods, our analysis will focus closely on the popular \emph{gradient-based} methods, Saliency Map~\citep{simonyan2013deep}, Integrated Gradient~\citep{sundararajan2017axiomatic} and Smooth Gradient~\citep{smilkov2017smoothgrad}, shown in Defs~\ref{def:saliency-map}-\ref{def:smoothed-gradient}.

\begin{definition}[Saliency Map (SM)]\label{def:saliency-map}
    The \emph{Saliency Map} $g_{\text{S}}(\xvec)$ is given by
    $
        g_{\text{S}}(\xvec) := \frac{\partial f(\xvec)}{\partial\xvec} $.
\end{definition}

\begin{definition}[Integrated Gradient (IG)]\label{def:integrated-gradient}
    Given a baseline input $\xvec_b$, the \emph{Integrated Gradient} $g_{\text{IG}}(\xvec ; \xvec_b)$ is given by
    $
        g_{\text{IG}}(\xvec; \xvec_b) := (\xvec-\xvec_b) \int^1_0 \frac{\partial f((\xvec-\xvec_b)t + \xvec_b)}{\partial\xvec} dt
        $.
\end{definition}

\begin{definition}[Smooth Gradient (SG)]\label{def:smoothed-gradient}
    Given a zero-centered Gaussian distribution $\mathcal{N}$ with a standard deviation $\sigma$, the \emph{Smooth Gradient} $g_{\text{SG}}(\xvec; \sigma)$ is given by
    $
        g_{\text{SG}}(\xvec; \sigma) :=  \mathbb{E}_{\boldsymbol{\epsilon}\sim\mathcal{N}(\mathbf{0}, \sigma^2I)} \frac{\partial f(\boldsymbol{\alpha} + \boldsymbol{\epsilon})}{\partial\xvec}
        $.
\end{definition}

Besides, we will also include results from DeepLIFT~\citep{shrikumar2017learning} and \texttt{grad} $\times$ \texttt{input} (element-wise multiplication between Saliency Map and the input)~\citep{simonyan2013deep} in our empirical evaluation. 
As we show in Section~\ref{sec:method-expl-locally-inear-models}, Defs~\ref{def:saliency-map}-\ref{def:smoothed-gradient} satisfy axioms that relate to the \emph{local linearity} of ReLU networks, and in the case of randomized smoothing~\citep{Cohen2019CertifiedAR}, their robustness to input perturbations.
We further discuss these methods relative to others in Sec.~\ref{sec:related-work}.

\noindent\textbf{Robustness.} Two relevant concepts about adversarial robustness will be used in this paper: \emph{prediction robustness} that the model's output label remains unchanged within a particular $\ell_p$ norm ball and \emph{attribution robustness} that the feature attributions are similar within the same ball. Recent work has identified the model's Lipschitz continuity as a bridge between these two concepts~\citep{NEURIPS2020_9d94c898} and some loss functions in achieving \emph{prediction robustness} also bring \emph{attribution robustness} ~\citep{chalasani2020concise}. We refer to \emph{robustness} as \emph{prediction robustness} if not otherwise noted.

%% file: 03_Method.tex

\section{Explainability, Decision Boundaries, and Robustness}\label{sec:method}
In this section, we begin by discussing the role of decision boundaries in constructing explanations of model behavior via feature attributions.
We first illustrate the key relationships in the simpler case of linear models, which contain exactly one boundary, and then generalize to piecewise-linear classifiers as they are embodied by deep ReLU networks.
We then show how local robustness causes attribution methods to align more closely with nearby decision boundaries, leading to explanations that better reflect these relationships.


\subsection{Attributions for linear models}\label{sec:method-expl-linear-models}

\begin{figure*}[!t]
    \centering
    \begin{subfigure}[b]{0.3\textwidth}
        \centering
        \includegraphics[width=\textwidth]{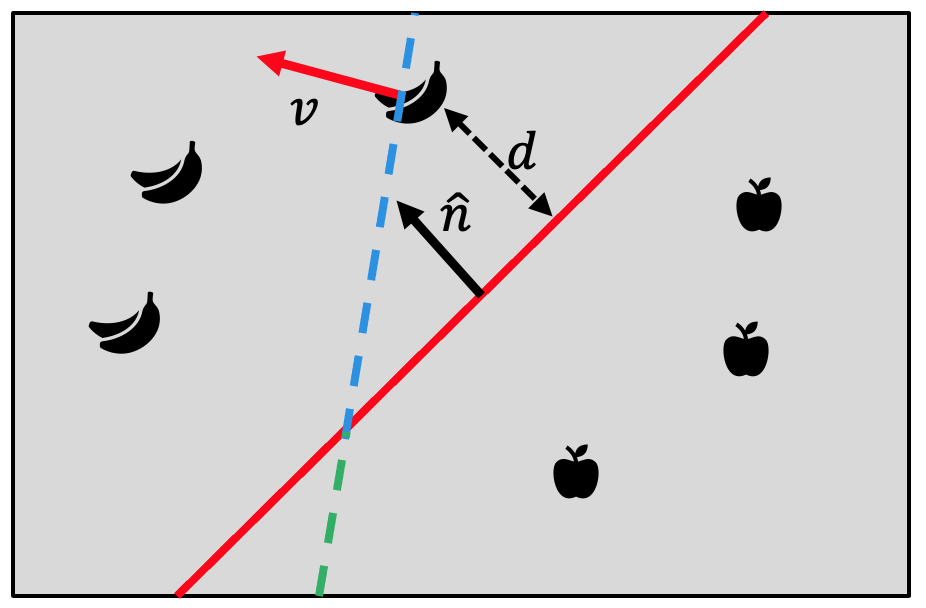}
        \caption{}
        \label{fig:linear-model}
    \end{subfigure}
    \hfill
    \begin{subfigure}[b]{0.3\textwidth}
        \centering
        \includegraphics[width=\textwidth]{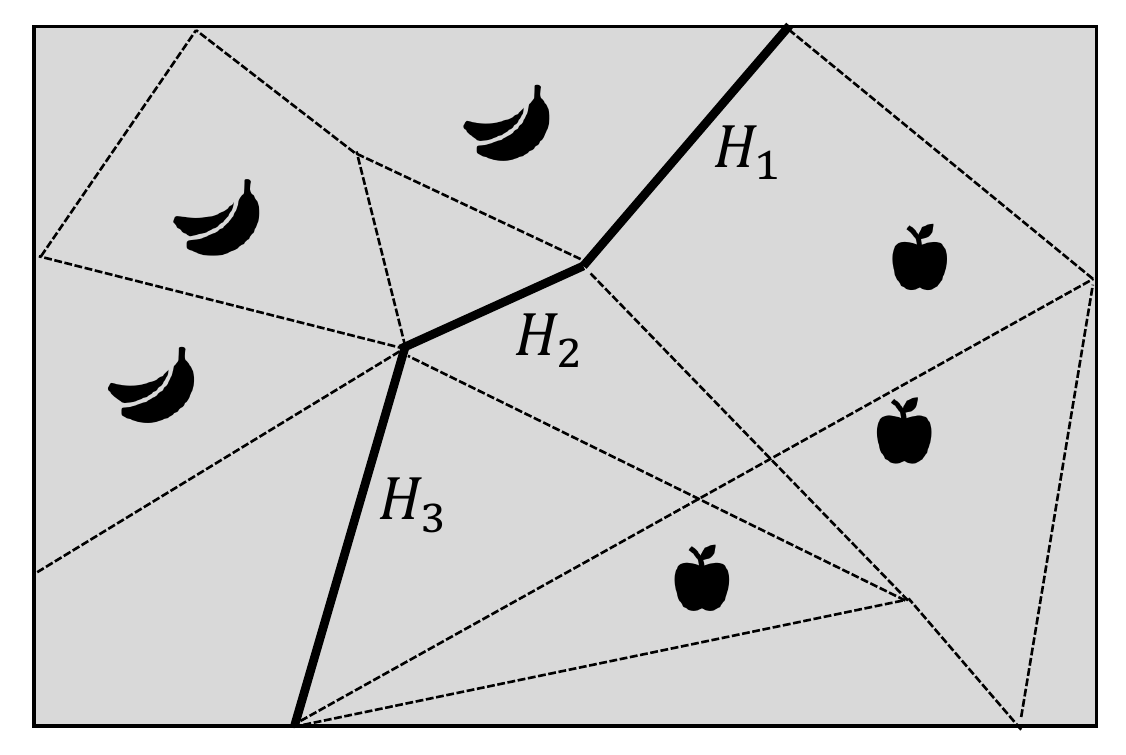}
        \caption{}
        \label{fig:local-linear-region}
    \end{subfigure}
    \hfill
    \begin{subfigure}[b]{0.3\textwidth}
        \centering
        \includegraphics[width=\textwidth]{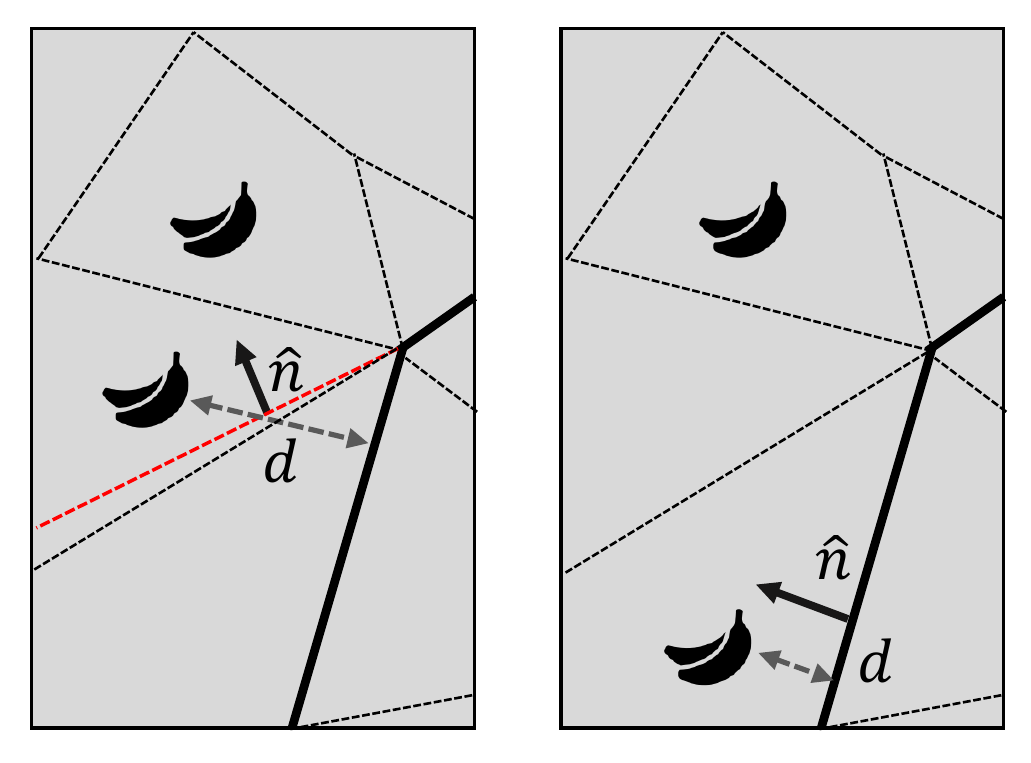}
        \caption{}
        \label{fig:projection-on-boundary}
    \end{subfigure}
       \caption{Different classifiers that partition the space into regions associated with \texttt{apple} or \texttt{banana}. (a) A linear classifier where $\hat{\mathbf{n}}$ is the only faithful explanations and $\mathbf{v}$ is not. (b) A deep network with ReLU activations. Solid lines correspond to decision boundaries while dashed lines correspond to facets of activation regions. (c) Saliency map of the target instance may be normal to the closest decision boundary (right) or normal to the prolongation of other local boundaries (left).}
       \label{fig:local-linear-decision-boundary}
\end{figure*}

Consider a binary classifier $C(\xvec) = \text{sign}(\wvec^\top\xvec + \bvec)$ that predicts a label $\{-1, 1\}$ (ignoring ``tie'' cases where $C(\xvec)=0$, which can be broken arbitrarily). 
In its feature space, $C(\xvec)$ is a hyperplane $H$ that separates the input space into two open half-spaces $S_1$ and $S_2$ (see Fig.~\ref{fig:linear-model}). 
Accordingly, the normal vector $\hat{\nvec}$ of the decision boundary is the only vector that faithfully explains the model's classification while other vectors, while they may describe directions that lead to positive changes in the model's output score, are not faithful in this sense (see $\mathbf{v}$ in Fig.~\ref{fig:linear-model} for an example).
In practice, to assign attributions for predictions made by $C$, SM, SG, and the integral part of IG (see Sec.~\ref{sec:background}) return a vector characterized by $\zvec = k_1\hat{\nvec} + k_2$~\citep{ancona2018towards}, where $k_1\ \ne 0$ and $k_2 \in \mathbb{R}$, regardless of the input $\xvec$ that is being explained. 
In other words, these methods all measure the importance of features by characterizing the model's decision boundary, and are equivalent up to the scale and position of $\hat{\nvec}$.

\subsection{Generalizing to piecewise-linear boundaries}\label{sec:method-expl-locally-inear-models}

In the case of a piecewise-linear model, such as a ReLU network, the decision boundaries comprise a collection of hyperplane segments that partition the feature space, as in $H_1, H_2$ and $H_3$ in the example shown in Figure~\ref{fig:local-linear-region}.
Because the boundary no longer has a single well-defined normal, one intuitive way to extend the relationship between boundaries and attributions developed in the previous section is to capture the normal vector of the \emph{closest} decision boundary to the input being explained. 
However, as we show in this section, the methods that succeeded in the case of linear models (SM, SG, and the integral part of IG) may in fact fail to return such attributions in the more general case of piecewise-linear models, but local robustness often remedies this problem.
We begin by reviewing key elements of the geometry of ReLU networks~\citep{Jordan2019ProvableCF}.

\noindent\textbf{ReLU activation polytopes.} 
For a neuron $u$ in a ReLU network $f(\xvec)$, we say that its status is \texttt{ON} if its pre-activation $u(\xvec) \geq 0$, otherwise it is \texttt{OFF}. 
We can associate an \emph{activation pattern} denoting the status of each neuron for any point $\xvec$ in the feature space, and a half-space $A_u$ to the activation constraint $u(\xvec) \geq 0$.
Thus, for any point $\xvec$ the intersection of the half-spaces corresponding to its activation pattern defines a polytope $P$ (see Fig.~\ref{fig:local-linear-region}), and within $P$ the network is a linear function such that $\forall \xvec \in P, f(\xvec)=\wvec_P^\top\xvec + b_P$, where the parameters $\wvec_p$ and $b_P$ can be computed by differentiation~\citep{fromherz20projections}. 
Each facet of $P$ (dashed lines in Fig.~\ref{fig:local-linear-region}) corresponds to a boundary that ``flips'' the status of its corresponding neuron.
Similar to activation constraints, decision boundaries are piecewise-linear because each decision boundary corresponds to a constraint $f_i(\xvec) \geq f_j(\xvec)$ for two classes $i, j$~\citep{fromherz20projections, Jordan2019ProvableCF}. 

\noindent\textbf{Gradients might fail.} 
Saliency maps, which we take to be simply the gradient of the model with respect to its input, can thus be seen as a way to project an input onto a decision boundary. 
That is, a saliency map is a vector that is normal to a nearby decision boundary segment. 
However, as others have noted, a saliency map is not always normal to any real boundary segment in the model's geometry~(see the left plot of Fig.~\ref{fig:projection-on-boundary}), because when the closest boundary segment is not within the activation polytope containing $\xvec$, the saliency map will instead be normal to the linear extension of some other hyperplane segment~\citep{fromherz20projections}. 
In fact, the fact that iterative gradient descent typically outperforms the Fast Gradient Sign Method~\citep{43405} as an attack demonstrates that this is often the case.

\noindent\textbf{When gradients succeed.} 
While saliency maps may not be the best approach in general for capturing information about nearby segments of the model's decision boundary, there are cases in which it serves as a good approximation.
Recent work has proposed using the Lipschitz continuity of an attribution method to characterize the difference between the attributions of an input $\xvec$ and its neighbors within a $\ell_p$ ball neighborhood (Def.~\ref{def:attribution-robustness})~\citep{NEURIPS2020_9d94c898}. 
This naturally leads to Proposition~\ref{theorem-robust-attributions}, which states that the difference between the saliency map at an input and the correct normal to the closest boundary segment is bounded by the distance to that segment.

	\begin{definition}[Attribution Robustness]\label{def:attribution-robustness}
		An attribution method $g(\xvec)$ is $(\lambda, \delta)$-locally robust at the evaluated point $\xvec$ if $\forall \xvec' \in B(\xvec, \delta), ||g(\xvec') - g(\xvec)|| \leq \lambda ||\xvec'-\xvec|| $.
	\end{definition}

\begin{proposition}\label{theorem-robust-attributions}
    Suppose that $f$ has a $(\lambda, \delta)$-robust saliency map $g_{\text{S}}$ at $\xvec$, $\xvec'$ is the closest point on the closest decision boundary segment to $\xvec$ and $||\xvec'-\xvec|| \leq \delta$, and that $\nvec$ is the normal vector of that boundary segment. 
    Then $||\nvec-g_{\text{S}}(\xvec)|| \leq \lambda ||\xvec-\xvec'||$. 
\end{proposition}

Proposition~\ref{theorem-robust-attributions} therefore provides the following insight: for networks that admit robust attributions~\citep{NEURIPS2019_172ef5a9, NEURIPS2020_9d94c898}, the saliency map is a good approximation to the boundary vector. 
As prior work has demonstrated the close correspondence between robust prediction and robust attributions~\citep{NEURIPS2020_9d94c898, chalasani2020concise}, this in turn suggests that explanations on robust models will more closely resemble boundary normals. 

As training robust models can be expensive, and may not come with guarantees of robustness, post-processing techniques like randomized smoothing~\citep{Cohen2019CertifiedAR}, have been proposed as an alternative.
\citet{Dombrowski2019ExplanationsCB} noted that models with softplus activations ($\mathbf{y} = 1/\beta \log(1+\exp{(\beta\xvec)})$) approximate smoothing, and in fact give an exact correspondence for single-layer networks.
Combining these insights, we arrive at Theorem~\ref{theorem-randomized-smoothing}, which suggests that the saliency map on a smoothed model approximates the closest boundary normal vector well; the similarity is inversely proportional to the standard deviation of the noise used to smooth the model.

\begin{theorem}\label{theorem-randomized-smoothing}
    Let $m(\xvec)$ be a one-layer ReLU network, and denote its smoothed counterpart under the Gaussian with standard deviation $\sigma$ as $m_\sigma(\xvec)$. 
    and $g(\xvec)$ be the saliency map for $m_\sigma(\xvec)$. 
    If $\xvec'$ is the closest adversarial example to $\xvec$, and $\forall \xvec''\in B(\xvec, ||\xvec-\xvec'||) . ||g(\xvec'')||\geq c$, then the following statement holds:
    $||g(\xvec) - g(\xvec') || \lessapprox \lambda
        $ where $\lambda \propto O(\frac{1}{\sigma})$. 
\end{theorem}


Theorem~\ref{theorem-randomized-smoothing} suggests that when randomized smoothing is used, the normal vector of the closest decision boundary segment and the saliency map are similar, and this similarity increases with the smoothness of the model's boundaries.
Because the saliency map for a smoothed model is equivalent to smooth gradient of its non-smooth counterpart~\citep{NEURIPS2020_9d94c898}, the smooth gradient is a better choice whenever computing the exact boundary is too expensive in a standard model. 
We provide empirical validation of this in Figure~\ref{fig:SG}.


\section{Boundary-Based Attribution}\label{sec:boundary-attributions}



Without the properties introduced by robust learning or randomized smoothing, the local gradient, i.e. saliency map, may not be a good approximation of decision boundaries. 
In this section, we build on the insights of our analysis to present a set of novel attribution methods that explicitly incorporate the normal vectors of nearby boundary segments.
Importantly, these attribution methods can be applied to models that are not necessarily robust, to derive explanations that capture many of the beneficial properties of explanations for robust models.

Using the normal vector of the closest decision boundary to explain a classifier naturally leads to Definition~\ref{def:boundary-based-saliency-map}, which defines attributions directly from the normal of the closest decision boundary.


\begin{definition}[Boundary-based Saliency Map (BSM)]\label{def:boundary-based-saliency-map}
    Given $f$ and an input $\xvec$, we define Boundary-based Saliency Map $B_{\text{S}}(\xvec)$ as follows:
    $
        B_{\text{S}}(\xvec) \defeq \partial f_c(\xvec')/\partial\xvec'
    $,
    where $\xvec'$ is the closest adversarial example to $\xvec$, i.e. $c = F(\xvec) \neq F(\xvec')$ and $\forall \xvec_m . ||\xvec_m-\xvec|| < ||\xvec'-\xvec|| \rightarrow F(\xvec) = F(\xvec_m)$.
\end{definition}


\paragraph{Incorporating More Boundaries.} 
The main limitation of using Definition~\ref{def:boundary-based-saliency-map} as a local explanation is obvious: the closest decision boundary only captures \emph{one} segment of the entire decision surface. 
Even in a small network, there will be numerous boundary segments in the vicinity of a relevant point. 
Taking inspiration from Integrated Gradients, Definition~\ref{def:boundary-based-integrated-gradient} proposes the Boundary-based Integrated Gradient (BIG) by aggregating the attributions along a line between the input and its closest boundary segment.

\begin{definition}[Boundary-based Integrated Gradient(BIG)]\label{def:boundary-based-integrated-gradient}
    Given $f$, Integrated Gradient $g_{\text{IG}}$ and an input $\xvec$, we define Boundary-based Integrated Gradient $B_{\text{S}}(\xvec)$ as follows:
    $
        B_{\text{IG}}(\xvec) := g_{\text{IG}}(\xvec;\xvec')
    $, where $\xvec$ is the nearest adversarial example to $\xvec$, i.e. $c = F(\xvec) \neq F(\xvec')$ and $\forall \xvec_m . ||\xvec_m-\xvec|| < ||\xvec'-\xvec|| \rightarrow F(\xvec) = F(\xvec_m)$.
\end{definition}

\paragraph{Geometric View of BIG.} 
BIG explores a linear path from the boundary point to the target point. 
Because points on this path are likely to traverse different activation polytopes, the gradient of intermediate points used to compute $g_{\text{IG}}$ are normals of linear extensions of their local boundaries.
As the input gradient is identical within a polytope $P_i$, the aggregate computed by BIG sums each gradient $\wvec_i$ along the path and weights it by the length of the path segment intersecting with $P_i$. 
In other words, one may view IG as an exploration of the model's global geometry that aggregates all boundaries from a fixed reference point, whereas BIG  explores the local geometry around $\xvec$. 
In the former case, the global exploration may reflect boundaries that are not particularly relevant to model's observed behavior at a point,
whereas the locality of BIG may aggregate boundaries that are more closely related (a visualization is shown in Fig.~\ref{fig:exmaple-bd}).

\noindent\textbf{Finding nearby boundaries.}
Finding the exact closest boundary segment is identical to the problem of certifying local robustness~\citep{fromherz20projections, Jordan2019ProvableCF, Kolter2018ProvableDA, lee2020lipschitz, leino2021globallyrobust, tjeng2018evaluating, Weng2018TowardsFC}, which is NP-hard for piecewise-linear models~\citep{sinha2020certifying}.
To efficiently find an approximation of the closest boundary segment, we leverage and ensemble techniques for generating adversarial examples, i.e. PGD~\citep{madry2018towards}, AutoPGD~\citep{croce2020reliable} and CW~\citep{Carlini2017TowardsET}, and use the closest one found given a time budget. 
The details of our implementation are discussed in Section~\ref{sec:evaluation}, where we show that this yields good results in practice.

%% file: 04_Evaluation.tex
\section{Evaluation}\label{sec:evaluation}

\begin{figure}[t]
    \centering
    \begin{subfigure}[b]{0.5\textwidth}
        \centering
       \begin{tabular}{ccccc} \toprule
       \small
           CIFAR10 & \multicolumn{2}{c}{standard} & \multicolumn{2}{c}{$\ell_2|0.5$}  \\ \midrule
           SM-BSM.  &\multicolumn{2}{c}{59.96}& \multicolumn{2}{c}{1.23}\\
           IG-AGI  & \multicolumn{2}{c}{28.20}&  \multicolumn{2}{c}{1.43}\\
           IG-BIG  & \multicolumn{2}{c}{31.22} & \multicolumn{2}{c}{2.73}\\ \midrule
           ImageNet & {standard} & {$\ell_2|3.0$} & {$\ell_{\infty}|\frac{4}{255}$}& {$\ell_{\infty}|\frac{8}{255}$}  \\ \midrule
           SM-BSM  &8.48& 0.41 & 2.25& 1.61\\
           IG-AGI  &13.52&  0.36&  1.19 & 0.86\\
           IG-BIG  & 17.07& 0.69 & 1.74& 1.45\\ \bottomrule
       \end{tabular}
        \caption{}
        \label{table:robust-difference}
    \end{subfigure}
    \hfill
    \begin{subfigure}[b]{0.45\textwidth}
       \centering
         \begin{tabular}{c|cccc} \toprule
         \small
             Corr. &Loc. & EG & PP & Con.\\ \midrule
             \textbf{SM}-BSM  &0.40& 0.46& -0.19&0.07\\
             \textbf{IG}-AGI  &0.24 &  0.25&  0.05 &-0.03\\
             \textbf{IG}-BIG  & 0.35& 0.30 & 0.20&-0.03\\ \bottomrule
         \end{tabular}
      \vspace{20pt}
     \caption{}
     \label{table:correlation}
      \end{subfigure}
       \caption{(a): $\ell_2$ differences between SM, IG and their boundary variants for robust models. The heading of each column reports the respective training epsilon and the corresponding $\ell_p$ norm constraint; Appendix~\ref{sec:appendix-boxplot} reports the corresponding boxplot. (b): Linear correlation coefficients between the alignment of SM and IG with nearby boundary vectors, and the localization metrics. For each row starting with $\mathbf{X}$-$Y$, the alignment is defined as $-||\mathbf{X}-Y||$. For each column, the localization results are measured with approach in bold font, a.k.a $\mathbf{X}$.}
       \label{fig:three graphs}
\end{figure}

\begin{table}[t]
    \centering
\begin{tabular}{cc|cccccccc} 
    Model &Metrics & {BIG} & {BSM} & {AGI} & {SM} & {GTI} & {SG} & {IG} & {DeepLIFT} \\
    \toprule
    &Loc.  &{\textbf{0.38}}& 0.33& 0.33& 0.33& 0.35& 0.34& 0.34& 0.34 \\
    standard&EG  & 0.54&  0.47&  0.48&  0.47&  0.46&  {\textbf{0.55}}&  0.5&  0.49 \\
    &PP  & {\textbf{0.87}}& 0.50& 0.58& 0.50& 0.50& 0.50& 0.51& 0.53  \\
    &Con.  &\textbf{4.35}& 3.88& 4.01& 3.92& 3.94& 4.06& 3.97& 3.93  \\  \midrule
    &Loc.  &{\textbf{0.39}}& 0.33& {\textbf{0.39}}& 0.33& 0.33& 0.34& 0.33& 0.33 \\
    $\ell_2|3.0$&EG  & {\textbf{0.74}}&  0.6&0.64& 0.6& 0.63& 0.62& 0.65& 0.64\\
    &PP  & {\textbf{0.92}}& 0.50& 0.88& 0.50& 0.55& 0.51& 0.65& 0.77  \\ 
    &Con.  & \textbf{5.03}& 4.12& 4.32&4.10 & 4.25 & 4.23& 4.37& 4.34 \\ 
    \bottomrule
\end{tabular}

    \caption{Results of several attribution methods over 1500 images of IamgeNet using a standard and robust ResNet50 (training $\epsilon$ is reported in the first column). BIG: Boundary-based Integrated Gradient. BSM: Boundary-based Saliency Map. AGI: Adversarial Gradient Integration. SM: Saliency Map. GTI: \texttt{grad}$\times$\texttt{input}. SG: Smoothed Gradient. IG: Integrated Gradient. See Appendix~\ref{appendix-visualizations-for-big} for the corresponding boxplot.}
    \label{tab:localization}
\end{table}


In this section, we first validate that the attribution vectors are more aligned to normal vectors of nearby boundaries in robust models(Fig.~\ref{table:robust-difference}). We secondly show that boundary-based attributions provide more ``accurate'' explanations -- attributions highlight features that are actually relevant to the label -- both visually (Fig.~\ref{fig:demo} and~\ref{fig:demo-robust}) and quantitatively (Table~\ref{tab:localization}). Finally, we show that in a standard model, whenever attributions more align with the boundary attributions, they are more ``accurate''. 

\noindent\textbf{General Setup.} We conduct experiments over two data distributions, ImageNet~\citep{ILSVRC15} and CIFAR-10~\citep{cifar}. For ImageNet, we choose 1500 correctly-classified images from ImageNette~\citep{imagenette}, a subset of ImageNet, with bounding box area less than 80\% of the original source image. For CIFAR-10, We use 5000 correctly-classified images. All standard and robust deep classifiers are ResNet50. All weights are pretrained and publicly available~\citep{robustness}. Implementation details of the boundary search (by ensembling the results of PGD, CW and AutoPGD) and the hyperparameters used in our experiments, are included in Appendix~\ref{appendix-experiment-setup-bounding-box}.

\subsection{Robustness $\rightarrow$ Boundary Alignment}\label{sec:eval-bounding-box}
In this subsection, we show that SM and IG better align with the normal vectors of the decision boundaries in robust models. For SM, we use BSM as the normal vectors of the nearest decision boundaries and measure the alignment by the $\ell_2$ distance between SM and BSM following Proposition~\ref{theorem-robust-attributions}. For IG, we use BIG as the aggregated normal vectors of all nearby boundaries because IG also incorporates more boundary vectors. Recently, \cite{pan2021explaining} also provides Adversarial Gradient Integral (AGI) as an alternative way of incorporating the boundary normal vectors into IG. We first use both BIG and AGI to measure how well IG aligns with boundary normals and later compare them in Sec.~\ref{sec:eval-visualization}, followed by a formal discussion in Sec.~\ref{sec:related-work}. 

Aggregated results for standard models and robust models are shown in Fig.~\ref{table:robust-difference}. It shows that adversarial training with bigger $\epsilon$ encourages a smaller difference between the difference between attributions and their boundary variants. Particularly, using $\ell_2$ norm and setting $\epsilon=3.0$ are most effective for ImageNet compared to $\ell_\infty$ norm bound. One possible explanation is that the $\ell_2$ space is special because training with $\ell_\infty$ bound may encourage the gradient to be more Lipschitz in $\ell_1$ because of the duality between the Lipschitzness and the gradient norm, whereas $\ell_2$ is its own dual.

\subsection{Boundary Attribution $\rightarrow$ Better Localization}\label{sec:eval-visualization}
In this subsection, we show boundary attributions (BSM, BIG and AGI) better localize relevant features. Besides SM, IG and SG, we also focus on other baseline methods including \texttt{Grad} $\times$ \texttt{Input} (GTI)~\citep{simonyan2013deep} and DeepLIFT (rescale rule only)~\citep{shrikumar2017learning} that are reported to be more faithful than other related methods~\citep{adebayo2018sanity, adebayo2020debugging}. 

In an image classification task where ground-truth bounding boxes are given, we consider features within a bounding box as more relevant to the label assigned to the image. Our evaluation is performed over ImageNet only because no bounding box is provided for CIFAR-10 data. The metrics used for our evaluation are: 1) \textbf{Localization (Loc.)}~\citep{chattopadhyay2017grad} evaluates the intersection of areas with the bounding box and pixels with positive attributions; 2) \textbf{Energy Game (EG)}~\citep{wang2020score} instead evaluates computes the portion of attribute scores within the bounding box. 
While these two metrics are common in the literature, we propose the following additional metrics: 3)\textbf{Positive Percentage (PP)} evaluates the portion of positive attributions in the bounding box because a naive assumption is all features within bounding boxes are relevant to the label (we will revisit this assumption in Sec.~\ref{sec:application}); and 4) \textbf{Concentration (Con.)} sums the absolute value of attribution scores over the distance between the ``mass" center of attributions and each pixel within the bounding box. 
Higher \textbf{Loc.}, \textbf{EG},  \textbf{PP} and \textbf{Con.} are better results. We provide formal details for the above metrics in Appendix~\ref{appendix-experiment-metrics}. 

We show the average scores for ResNet50 models in Table~\ref{tab:localization} where the corresponding boxplots can be found in Appendix~\ref{sec:appendix-boxplot}. BIG is noticeably better than other methods on Loc. EG, PP and Con. scores for both robust and standard models and matches the performance of SG on EG for a standard model. Notice that BSM is not significantly better than others in a standard model, which confirms our motivation of BIG -- that we need to incorporate more nearby boundaries because a single boundary may not be sufficient to capture the relevant features. 

We also measure the correlation between the alignment of SM and BSM with boundary normals and the localization abilities, respectively. For SM, we use BSM to represent the normal vectors of the boundary. For IG, we use AGI and BIG. For each pair $\mathbf{X}$-$Y$ in $\{\text{\textbf{SM}-BSM, \textbf{IG}-AGI, \textbf{IG}-BIG}\}$, we measure the empirical correlation coefficient between $-||\mathbf{X}-Y||_2$ and the localization scores of $\mathbf{X}$ in a standard ResNet50 and the result is shown in Fig.~\ref{table:correlation}. Our results suggest that when the attribution methods better align with their boundary variants, they can better localize the relevant features in terms of the Loc. and EG. However, PP and Con. have weak and even negative correlations. One possible explanation is that the high PP and Con. of BIG and AGI compared to IG (as shown in Table~\ref{tab:localization}) may also come from the choice of the reference points. Namely, compared to a zero vector, a reference point on the decision boundary may better filter out noisy features.

\begin{figure}[!t]
    \centering
    \includegraphics[width=\textwidth]{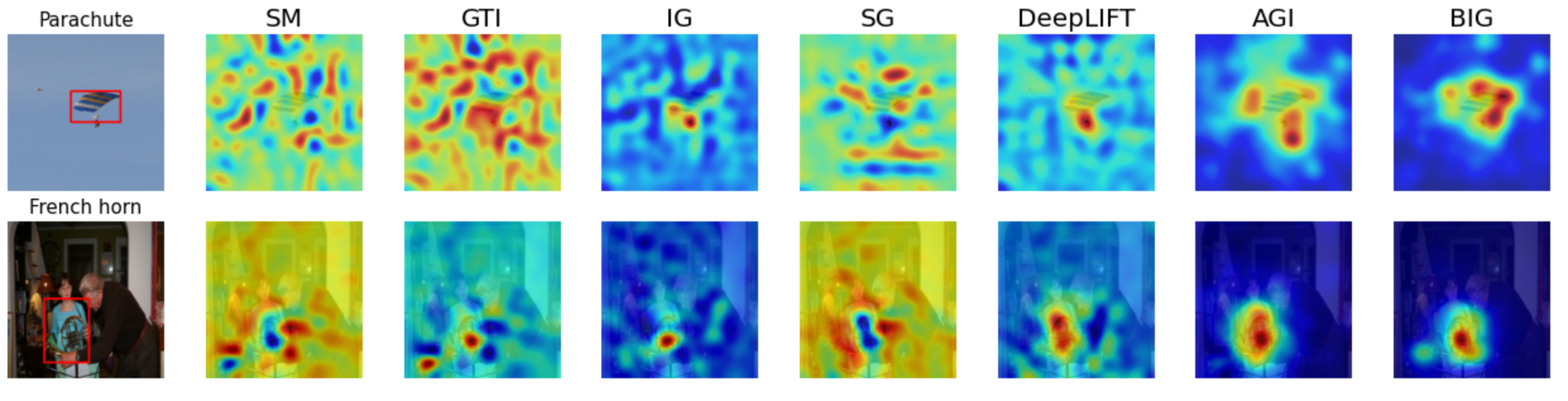}
    \caption{Visualizations of attributions for two examples classified by a standard ResNet50.}\label{fig:demo}
\end{figure}

\begin{figure}[!t]
    \centering
    \includegraphics[width=\textwidth]{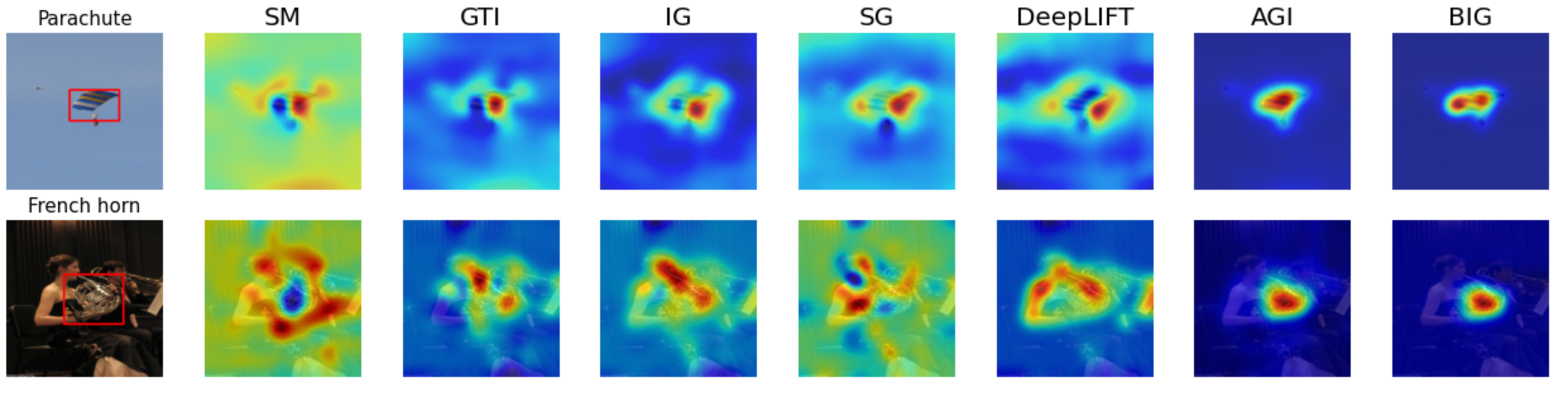}
    \caption{Visualizations of attributions for two examples classified by a robust ResNet50 ($\ell_2|3.0$). The second example from Fig.~\ref{fig:demo} is not correctly classified so we replace it with another image.}\label{fig:demo-robust}
\end{figure}

We end our evaluations by visually comparing the proposed method, BIG, against all other attribution methods for the standard ResNet50 in Fig.~\ref{fig:demo} and for the robust ResNet50 in Fig.~\ref{fig:demo-robust}, which demonstrates that BIG can easily and efficiently localize features that are relevant to the prediction. More visualizaitons can be found in the Appendix~\ref{appendix-visualizations-for-big}. 

\noindent\textbf{Summary.} Taken together, we close the loop and empirical show that standard attributions in robust models are visually more interpretable because they better capture the nearby decision boundaries. Thefore, the final take-away from our analitical and empirical results is if more resources are devoted to training robust models, effectively identical explanations can be obtained using much less costly standard gradient-based methods, i.e. IG.


%% file: 05_Application.tex
\section{Discussion}\label{sec:application}

\newcommand{\cmark}{\ding{51}}%
\newcommand{\xmark}{\ding{55}}%

\begin{figure}[t]
     \centering
     \begin{subfigure}[b]{0.45\textwidth}
      \centering
        \begin{tabular}{ccccc}
    \toprule
    {Properties} & {black IG} & {AGI} & {BIG}\\ \midrule
    {Boundary-based}  & \xmark &\cmark & \cmark \\
    {Boundary Search}  & N/A  & PGD & Any  \\ 
    {Geometry}  & Global  & Local & Local  \\ 
    {Symmetry}  &  \cmark &\xmark  &\cmark    \\
    {Completeness} &\cmark & \cmark & \cmark   \\
     \bottomrule
        \end{tabular}
        \vspace{20pt}
        \caption{}
        \label{fig:comparing with agi}
     \end{subfigure}
     \hfill
     \begin{subfigure}[b]{0.45\textwidth}
         \centering
         \includegraphics[width=\textwidth]{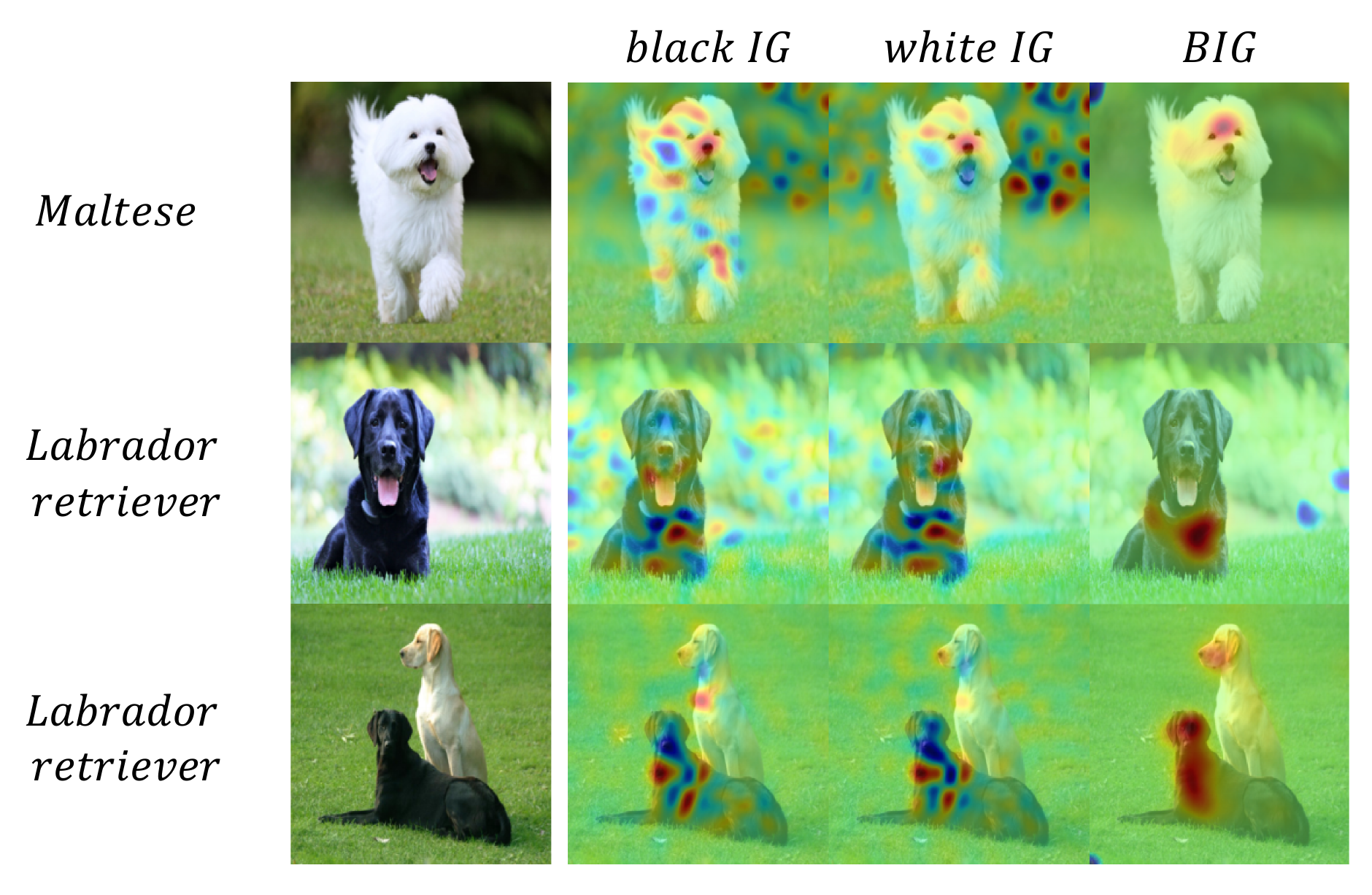}
         \caption{}
         \label{fig:baseline-demo}
     \end{subfigure}
    \caption{(a): Qualitative comparisons between IG with black baseline, BIG and AGI. BIG can use any boundary search approaches or an ensemble of them while AGI uses PGD only. AGI fails to meet the \emph{symmetry} axiom~\citep{sundararajan2017axiomatic} where BIG satisfies all axioms that IG satisfies, i.e. \emph{completeness}. (b): Comparisons of IG with black and white baselines with BIG. Predictions are shown in the first column.}
        \label{fig:comparisons}
\end{figure}

\noindent\textbf{Baseline Sensitivity.} It is naturally to treat BIG frees users from the baseline selection in explaining non-linear classifiers. Empirical evidence has shown that IG is sensitive to the baseline inputs~\citep{sturmfels2020visualizing}. We compare BIG with IG when using different baseline inputs, white or black images. We show an example in Fig~\ref{fig:baseline-demo}. For the first two images, when using the baseline input as the opposite color of the dog, more pixels on dogs receive non-zero attribution scores. Whereas backgrounds always receive more attribution scores when the baseline input has the same color as the dog. This is because $g_{\text{IG}}(\xvec)_i \propto (\xvec-\xvec_b)_i$ (see Def.~\ref{def:integrated-gradient}) that greater differences in the input feature and the baseline feature can lead to high attribution scores. The third example further questions the readers using different baselines in IG whether the network is using the white dog to predict \texttt{Labrador retriever}. We demonstrate that conflicts in IG caused by the sensitivity to the baseline selection can be resolved by BIG. BIG shows that black dog in the last row is more important for predicting \texttt{Labrador retriever} and this conclusion is further validated by our counterfactual experiment in Appendix~\ref{appendix-counterfactual-analysis-for-baseline}. Overall, the above discussion highlights that BIG is significantly better than IG in reducing the non-necessary sensitivity in the baseline selection. 

\noindent\textbf{Limitations.} We identify two limitations of the work. 1) Bounding boxes are not perfect ground-truth knowledge for attributions. In fact, we find a lot of examples where the bounding boxes either fail to capture all relevant objects or are too big to capture relevant features only. Fixing mislabeled bounding boxes still remain an open question and should benefit more expandability research in general. 2) Our analysis only targets on attributions that are based on end-to-end gradient computations. That is, we are not able to directly characterize the behavior of perturbation-based approaches, i.e. Mask~\citep{fong2017interpretable}, and activation-based approaches, i.e. GradCAM~\citep{selvaraju2017grad} and Feature Visualization~\citep{olah2017feature}. 

%% file: 06_RelatedWork.tex
\section{Related Work}\label{sec:related-work}

\cite{NEURIPS2019_e2c420d9} shows an alternative way of explaining why robust models are more interpretable by showing robust models usually learn robust and relevant features, whereas our work serves as a geometrical explanation to the same empirical findings in using attributions to explain deep models. Our analysis suggests we need to capture decision boundaries in order to better explain classifiers, whereas a similar line of work, AGI~\citep{pan2021explaining} that also involves computations of adversarial examples is motivated to find a non-linear path that is linear in the representation space instead of the input space compared to IG. Therefore, AGI uses PGD to find the adversarial example and aggregates gradients on the non-linear path generated by the PGD search. We notice that the trajectory of PGD search is usually extremely non-linear, complex and does not guarantee to return closer adversarial examples without CW or AutoPGD (see comparisons between boundary search approaches in Table~\ref{appendix-experiment-setup-bounding-box}). We understand that finding the exact closest decision boundary is not feasible, but our empirical results suggest that the linear path (BIG) returns visually sharp and quantitative better results in localizing relevant features. Besides, a non-linear path should cause AGI fail to meet the \emph{symmetry} axiom~\citep{sundararajan2017axiomatic} (see Appendix~\ref{sec:appendix-symmetry} for an example of the importance of \emph{symmetry} for attributions). We further summarize the commons and differences in Table~\ref{fig:comparing with agi}. 

In the evaluation of the proposed methods, we choose metrics related to bounding box over other metrics because for classification we are interested in whether the network associate relevant features with the label while other metrics~\citep{adebayo2018sanity, ancona2017better, samek2016evaluating, wang2020interpreting, yeh2019fidelity}, e.g. infidelity~\citep{yeh2019fidelity}, mainly evaluates whether output scores are faithfully attributed to each feature. Our idea of incorporating boundaries into explanations may generalize to other score attribution methods, e.g. Distributional Influence~\citep{leino2018influence} and DeepLIFT~\citep{shrikumar2017learning}. The idea of using boundaries in the explanation has also been explored by T-CAV~\citep{Kim2018InterpretabilityBF}, where a linear decision boundary is learned for the internal activations and associated with their proposed notion of \emph{concept}. 

When viewing our work as using nearby boundaries as a way of exploring the local geometry of the model's output surface, a related line of work is NeighborhoodSHAP~\citep{Ghalebikesabi2021OnLO}, a local version of SHAP~\citep{lundberg2017unified}. When viewing our as a different use of adversarial examples, some other work focuses on counterfactual examples (semantically meaningful adversarial examples) on the data manifold~\citep{Chang2019ExplainingIC, Dhurandhar2018ExplanationsBO, pmlr-v97-goyal19a}.

%% file: 07_Conlusion.tex
\section{Conclusion}\label{sec:conclusion}

In summary, we rethink the target question an explanation should answer for a classification task, the important features that the classifier uses to place the input into a specific side of the decision boundary. We find the answer to our question relates to the normal vectors of decision boundaries in the neighborhood and propose BSM and BIG as boundary attribution approaches. Empirical evaluations on STOA classifiers validate that our approaches provide more concentrated, sharper and more accurate explanations than existing approaches. Our idea of leveraging boundaries to explain classifiers connects explanations with the adversarial robustness and help to encourage the community to improve model quality for explanation quality. 

%% file: 08_Appendix.tex
\newpage
	\section{Theorems and Proofs}\label{appendix:proofs}
	
	\subsection{Proof of Proposition~\ref{theorem-robust-attributions}}\label{appendix-proof-theorem-2}
	
	\noindent\textbf{Proposition~\ref{theorem-robust-attributions}} \textit{
    Suppose that $f$ has a $(\lambda, \delta)$-robust saliency map $g_{\text{S}}$ at $\xvec$, $\xvec'$ is the closest point on the closest decision boundary segment to $\xvec$ and $||\xvec'-\xvec|| \leq \delta$, and that $\nvec$ is the normal vector of that boundary segment. Then $||\nvec-g_{\text{S}}(\xvec)|| \leq \lambda ||\xvec-\xvec'||$. 
	}
	
	To compute $\nvec$ can be efficiently computed by taking the derivatice of the model's output w.r.t to the point that is on the decision boundary such that $\nvec = \frac{\partial f(\xvec')}{\partial \xvec'}$ and $\forall \xvec_m \in \mathbb{R}^d, F(\xvec_m) = F(\xvec)$ if $||\xvec_m - \xvec|| \leq ||\xvec'-\xvec||$.
	
	Because we assume $||\xvec - \xvec'|| \leq \delta$, and the model has $(\lambda, \delta)$-robust Saliency Map, then by Def.~\ref{def:attribution-robustness} we have
	$$||\nvec-g_{\text{S}}(\xvec)|| \leq \lambda ||\xvec-\xvec'||$$
	
	

	\subsection{Proof of Theorem~\ref{theorem-randomized-smoothing}}\label{appendix-proof-theorem-1}
	
	\noindent\textbf{Theorem \textcolor{red}{1}}
	\textit{Let $m(\xvec) = ReLU(\wvec^\top\xvec)$ be a one-layer network and when using randomized smoothing, we write $m_\sigma(\xvec)$. Let $g(\xvec)$ be the SM for $m_\sigma(\xvec)$ and suppose $\forall \xvec''\in B(\xvec, ||\xvec-\xvec'||), ||g(\xvec'')||\geq c$ where $\xvec'$ is the closest adversarial example, we have the following statement holds:}
	$
	||g(\xvec) - g(\xvec') || \lessapprox \lambda
	$ \textit{where} $\lambda \propto O(\frac{1}{\sigma})$. 
	
	Before we start the proof, we firstly introduce \emph{Randomized Smoothing} and its theorem that certify the robustness.
	
	\begin{definition}[Randomized Smoothing~\citep{Cohen2019CertifiedAR}]\label{def:randomized_smoothing}
		Suppose $F(\xvec) = \arg\max_c f_c(\xvec)$, the smoothed classifier $G(\xvec)$ is defined as 
		\begin{align}
			G(\xvec) := \arg\max_c \Pr{[F(\xvec + \epsilon) = c]} 
		\end{align} where $\epsilon \sim \mathcal{N}(\mathbf{0}, \sigma^2I )$
	\end{definition}
	
	\begin{theorem}[Theorem 1 from Cohen et al.~\citep{Cohen2019CertifiedAR}]\label{theorem:randomized smoothing}
		Suppose $F$ and $G$ are defined in Def.~\ref{def:randomized_smoothing}. For a target instance $\xvec$, suppose $\Pr[F(\xvec)=c_A]$ is lower-boundded by $p_A$ and $\max_{c\neq c_A} \Pr[F(\xvec)=c]$ is upper-boundded by $p_B$, then
		\begin{align}
			G(\xvec') = c_A \quad \forall \xvec' \in B(\xvec, R_\sigma) 
		\end{align} where \begin{align}
		R_\sigma=\frac{\sigma}{2}[\Phi^{-1}(p_A) - \Phi^{-1}(p_B)]\label{eq:R-sigma}
		\end{align} and $\Phi$ is the c.d.f of Gaussian. 
	\end{theorem}
	
	We secondly introduce a theorem that connects \emph{Randomized Smoothing} and Smoothed Gradient.
	
	\begin{theorem}[Proposition 1 from Want et al.~\citep{NEURIPS2020_9d94c898}]~\label{theorem:SG-and-RS}
		Suppose a model $f(\xvec)$ satisfies $|\max|f(\xvec)| < \infty$. For a Smoothed Gradient $g_{\text{SG}}(\xvec)$, we have
		\begin{align}
			g_{\text{SG}}(\xvec) = \frac{\partial (f \circledast q)(\xvec)}{\partial\xvec} 
		\end{align} where $q(\xvec) = \mathcal{N}(\mathbf{0}, \sigma^2 I)$ and $\circledast$ denotes the convolution operation.
	\end{theorem}
	
	Finally, we introduce two theorems that connects Smoothed Gradient with Softplus network. 
	
	\begin{theorem}[Thoerem 1 from Dombrowski et al.~\citep{Dombrowski2019ExplanationsCB}]\label{theorem:range-of-SM-softpus}
		Suppose $m(\xvec)$ is a feed-forward network with softplus-$\beta$ activation and $||\frac{\partial m(\xvec)}{\partial \xvec}|| \geq c$ if $\xvec \in B(\xvec, \epsilon)$, then 
		\begin{align}
			||\frac{\partial m(\xvec)}{\partial \xvec} - \frac{\partial m(\xvec')}{\partial \xvec'}|| \leq \beta C d_g(\xvec, \xvec') 
		\end{align} where $$C = \frac{1}{c} \sum_i ||W^{(L)}||_F||W^{(L-1)}||_F...||W^{(i)}||^2_F...||W^{(1)}||^2_F$$ $W^{(i)}$ is the weight for layer $i$ and $d_g(\xvec, \xvec')$ is the \emph{geodesic distance} (which in our case is just the $\ell_2$ distance).
	\end{theorem}
	
	\begin{theorem}[Thoerem 2 from Dombrowski et al.~\citep{Dombrowski2019ExplanationsCB}]\label{theorem:d2}
		Denote a one-layer ReLU network as $f(\xvec) = ReLU(\wvec^\top\xvec)$ and a one-layer softplus network as $m_\beta(\xvec) = softplus_\beta(\wvec^\top\xvec)$, then the following statement holds:
		\begin{align}
			\mathbb{E}_{\epsilon_i \sim p_\beta}[\frac{\partial f(\xvec+\epsilon)}{\partial \xvec}] = \frac{\partial m_{\frac{\beta}{||\wvec||}}(\xvec)}{\partial \xvec} 
		\end{align} where $p_\beta(\epsilon_i) = \frac{\beta}{(\exp{(\beta \epsilon_i/2)} + \exp{(-\beta\epsilon_i / 2)})^2}$
	\end{theorem}
	
	We now begin our proof for Theorem~\ref{theorem-randomized-smoothing}.
	
	\noindent\textit{Proof:}
	
	\textit{
		Given a one-layer ReLU network $m(\xvec) = ReLU(\wvec^\top\xvec)$ that takes an input $\xvec \in \mathbb{R}^d$ and outputs the logit score for the class of interest. WLOG we assume $\wvec \in \mathbb{R}^d$ is the $j$-th column of the complete weight matrix $W \in \mathbb{R}^{d\times n}$, where $j$ is the class of interest. With Theorem~\ref{theorem:d2} we know that
		\begin{align}
			\mathbb{E}_{\epsilon_i \sim p_\beta}[\frac{\partial m(\xvec+\epsilon)}{\partial \xvec}] = \frac{\partial m_{\frac{\beta}{||\wvec||}}(\xvec)}{\partial \xvec}\label{eq: softplus-relu} 
		\end{align}
		Dombrowski et al.~\citep{Dombrowski2019ExplanationsCB} points out that the random distribution $p_\beta(\epsilon_i) = \frac{\beta}{(\exp{(\beta \epsilon_i/2)} + \exp{(-\beta \epsilon_i / 2)})^2}$ closely resembles a normal distribution with a standard deviation 
		\begin{align}
			\sigma = \sqrt{\log(2)\frac{\sqrt{(2\pi)}}{\beta}}\label{eq:sigma-beta} 
		\end{align}
		Therefore, we have the following relation  
		\begin{align}
			\mathbb{E}_{\epsilon_i \sim p_\beta}[\frac{\partial m(\xvec+\epsilon)}{\partial \xvec}] \approx \mathbb{E}_{\epsilon\sim \mathcal{N}(0, \sigma^2_\beta I)}[\frac{\partial m(\xvec+\epsilon)}{\partial \xvec}]\label{eq: softplus-SG} 
		\end{align} 
		The LHS of the above equation is Smoothed Gradient, or equaivelently, it is the Saliency Map of a smoothed classifier $m_{\sigma}$ due to Theorem~\ref{theorem:SG-and-RS}. Eq~\ref{eq: softplus-relu} and~\ref{eq: softplus-SG} show that we can analyze randomized smoothing with the tool of an intermeidate softplus network $m_{\beta/||\wvec||}$.
		\begin{align}
			\frac{\partial m_{\sigma}(\xvec)}{\partial \xvec} = \mathbb{E}_{\epsilon\sim \mathcal{N}(0, \sigma^2_\beta I)}[\frac{\partial m(\xvec+\epsilon)}{\partial \xvec}] \approx \frac{\partial m_{\frac{\beta}{||\wvec||}}(\xvec)}{\partial \xvec} 
		\end{align} and we denote $g(\xvec) := \frac{\partial m_{\sigma}(\xvec)}{\partial \xvec}$. Now consider Theorem~\ref{theorem:range-of-SM-softpus}, we have 
		\begin{align}
			||g(\xvec)-g(\xvec')|| & \approx ||\frac{\partial m_{\frac{\beta}{||\wvec||}}(\xvec)}{\partial \xvec}-\frac{\partial m_{\frac{\beta}{||\wvec||}}(\xvec')}{\partial \xvec'} || \\
			                       & \leq \frac{\beta}{c||\wvec||} ||W||^2_F ||\xvec-\xvec'||                                                                                             
		\end{align} where $\xvec'$ is the closest adversarial example. Because $m_{\sigma}$ is certified to be robust within the neighborhood $B(\xvec, R_{\sigma})$, therefore, the closest decision boundary is at least $R_{\sigma}$ distance away from the evaluated point $\xvec$. We then have
		\begin{align}
			||g(\xvec)-g(\xvec')|| \lessapprox \frac{\beta}{c||\wvec||} ||W||^2_F R_{\sigma} 
		\end{align} Now lets substitute $\beta$ and $R_{\sigma}$ with $\sigma$ using Eq.~\ref{eq:sigma-beta} and~\ref{eq:R-sigma}, we arrive at
		\begin{align}
			||g(\xvec)-g(\xvec')|| & \lessapprox \lambda 
		\end{align} where \begin{align}
		\lambda &= \frac{||W||^2_F}{c||\wvec||}\log(2)\frac{\sqrt{(2\pi)}}{\sigma^2}\frac{\sigma}{2}[\Phi^{-1}(p_A) - \Phi^{-1}(p_B)]\\
		&= \frac{1}{\sigma} \frac{\sqrt{(2\pi)}||W||^2_F}{2c||\wvec||}\log(2)[\Phi^{-1}(p_A) - \Phi^{-1}(p_B)]\\
		& \propto O(\frac{1}{\sigma})
		\end{align}
		We use $\lessapprox$ instead of $\leq$ because of the fact we approximate the randomnized smoothing with a similar distribution $p_\beta(\epsilon_i) = \frac{\beta}{(\exp{(\beta \epsilon_i/2)} + \exp{(-\beta \epsilon_i / 2)})^2}$. In fact, a more rigorous proof exists when considering $p_\beta$ other than Gaussian. We refer the readers who are interested in the content to Yang et al.~\citep{Yang2020RandomizedSO}.
	}
	
	\section{Experiment Details and Additional Results}
	
	\subsection{Metrics with Bounding Boxes}\label{appendix-experiment-metrics}
	
	We will use the following extra notations in this section. Let $X$, $Z$ and $U$  be a set of indices of all pixels, a set of indices of pixels with positive attributions, and a set of indices of pixels inside the bounding box for a target attribution map $g(\xvec)$. We denote the cardinality of a set $S$ as $|S|$.
	
	\paragraph{Localization (Loc.)}~\citep{chattopadhyay2017grad} evaluates the intersection of areas with the bounding box and pixels with positive attributions. 
	
	\begin{definition}[Localization]
		For a given attribution map $g(\xvec)$, the localization score (Loc.) is defined as 
		\begin{align}
			Loc := \frac{|Z \cap U|}{|U| + |Z\cap(X\setminus U)|} 
		\end{align}
	\end{definition}
	
	\paragraph{Energy Game (EG)}~\citep{wang2020score} instead evaluates computes the portion of attribute scores within the bounding box. 
	
	\begin{definition}[Energy Game]
		For a given attribution map $g(\xvec)$, the energy game EG is defined as
		\begin{align}
			EG := \frac{\sum_{i \in {Z \cap U}}g(\xvec)_i}{\sum_{i \in X}\max(g(\xvec)_i, 0)} 
		\end{align}
	\end{definition}
	
	\paragraph{Positive Percentage (PP)} evaluates the sum of positive attribute scores over the total (absolute value of) attribute scores within the bounding box. 
	
	\begin{definition}[Positive Percentage]
		Let $V$ be a set of indices pf all pixels with negative attribution scores, for a given attribution map $g(\xvec)$, the positive percentage PP is defined as
		\begin{align}
			PP := \frac{\sum_{i \in {Z \cap U}}g(\xvec)_i}{\sum_{i \in {Z \cap U}}g(\xvec)_i - \sum_{i \in {V \cap U}}g(\xvec)_i } 
		\end{align}
	\end{definition}
	
	\paragraph{Concentration (Con.)} evaluates the sum of weighted distances by the ``mass'' between the ``mass" center of attributions and each pixel within the bounding box. Notice that the computation of $c_x$ and $c_y$ can be computed with \texttt{scipy.ndimage.center\_of\_mass}. This definition encourages that pixels with high absolute value of attribution scores to be closer to the mass center.
	
	\begin{definition}[Concentration]
		For a given attribution map $g(\xvec)$, the concentration Con. is defined as follws
		\begin{align}
			Con. := \sum_{i \in U} \hat{g}(\xvec)_i / \sqrt{(i_x - c_x)^2 + (i_y - c_y)^2} 
		\end{align} where $\hat{g}$ is the normalized attribution map so that $\hat{g}_i = g_i / \sum_{i \in U} |g_i|$. $i_x, i_y$ are the coordinates of the pixel and \begin{align}
		c_x = \frac{\sum_{i \in  U} i_x \hat{g}(\xvec)_i}{\sum_{i \in U} \hat{g}(\xvec)_i}, c_y = \frac{\sum_{i \in U} i_y \hat{g}(\xvec)_i}{\sum_{i \in U} \hat{g}(\xvec)_i}
		\end{align}
	\end{definition}
	
	Besides metrics related to bounding boxes, there are other metrics in the literature used to evaluate attribution methods~\citep{adebayo2018sanity, ancona2017better, samek2016evaluating, wang2020interpreting, yeh2019fidelity}.
	We focus on metrics that use provided bounding boxes, as we believe that they offer a clear distinction between likely relevant features and irrelevant ones.

	\begin{figure*}[!t]
		\centering
		\includegraphics[width=\textwidth]{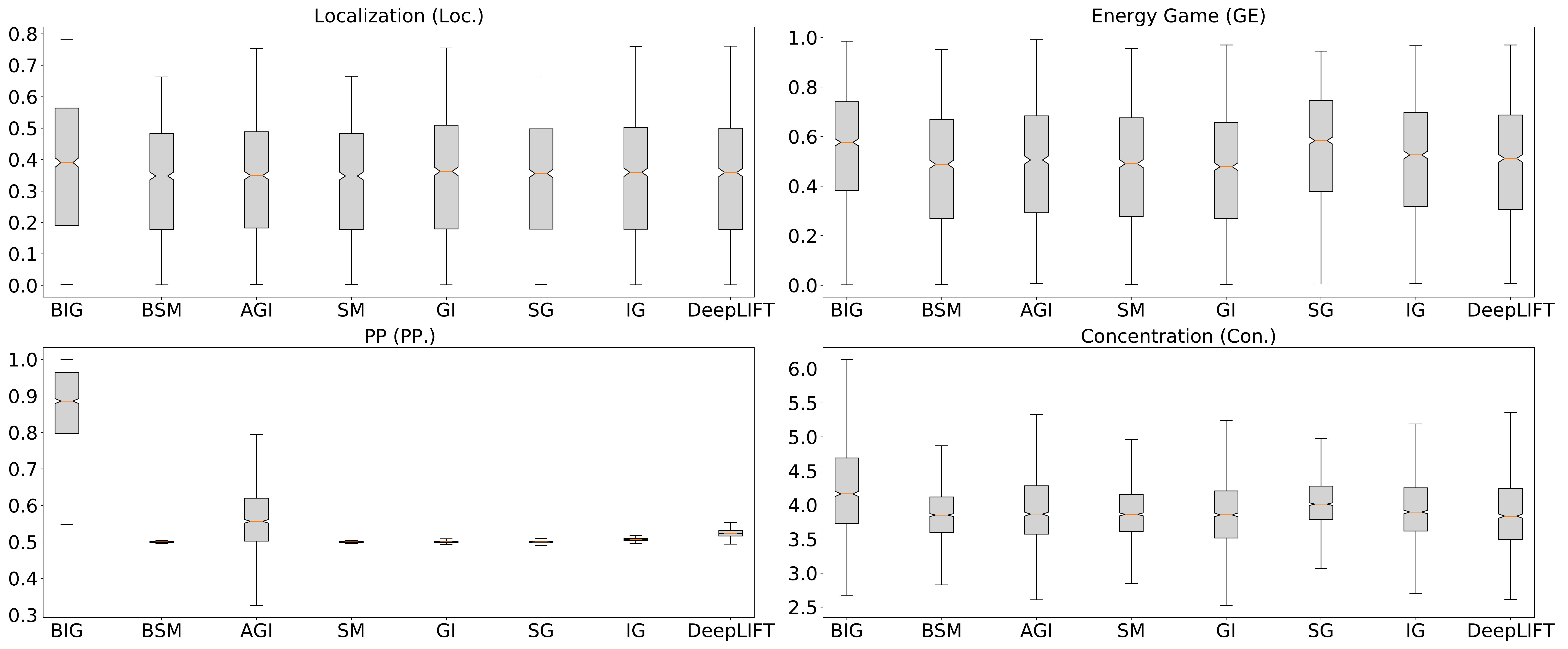}
		\caption{Localizaiton performance for attributions on a standard ResNet}\label{fig:more-attribution-comparison}
	\end{figure*}

	\begin{figure*}[!t]
		\centering
		\includegraphics[width=\textwidth]{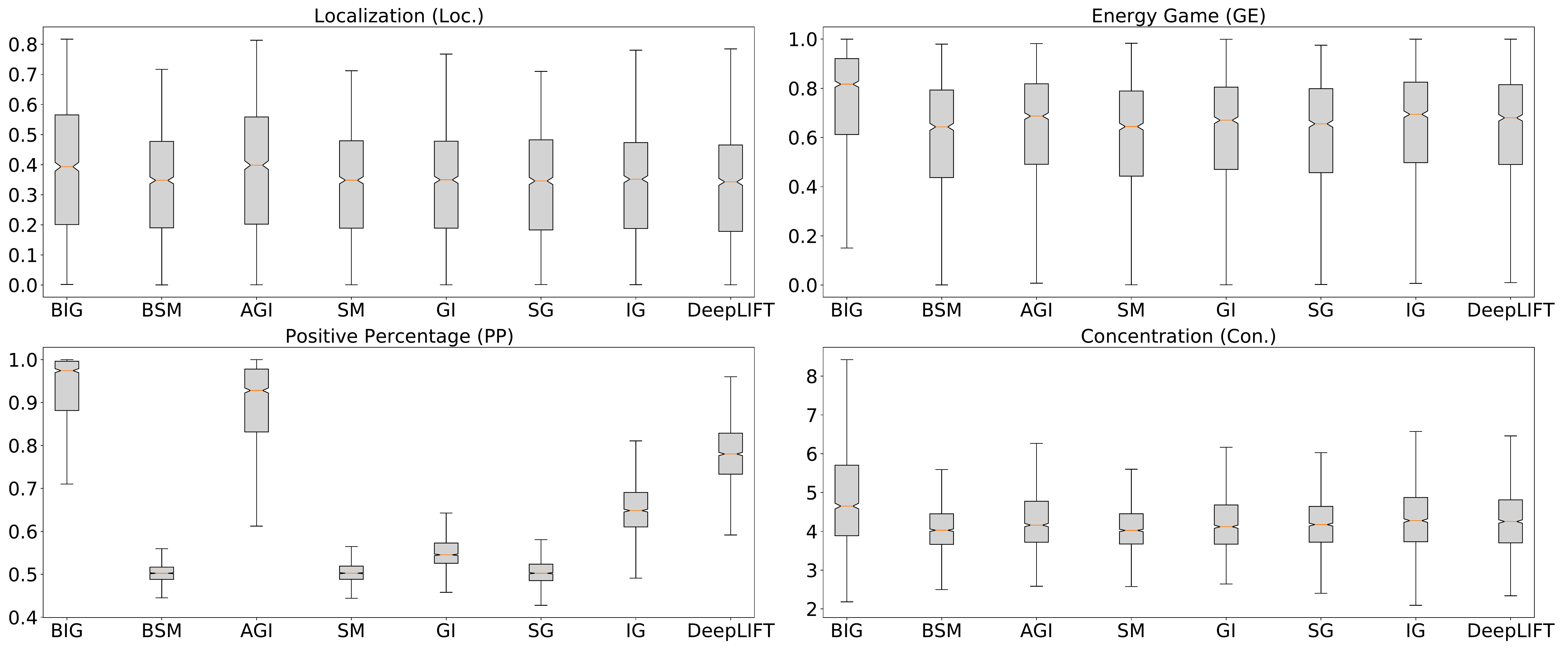}
		\caption{Localizaiton performance for attributions on a robust ResNet ($\ell_2|3.0$)}\label{fig:robust-more-attribution-comparison}
	\end{figure*}

	\subsection{Implementing Boundary Search}\label{appendix-experiment-setup-bounding-box}

	\begin{figure}[t]
		\centering
		\begin{subfigure}[b]{0.45\textwidth}
			\centering
			\small
			\begin{tabular} {|c|c|c|}
				\hline
				\textit{Pipeline} & \textit{Avg Distance} & \textit{Success Rate}\\
				\hline
				\multicolumn{3} {|c|}{\textbf{(ImageNet) Standard ResNet50}}\\
				\hline
				PGDs              & 0.549                  & 72.1\%                 \\
				~~~ + CW          & 0.548                  & 72.1\%              \\
				~~~~~~~~~~~~~~~~~ + AutoPGD          & 0.548                  & 72.1\%                    \\
				\hline
				\multicolumn{3} {|c|}{\textbf{(ImageNet) Robust ResNet50 ($\ell_2 | 3.0$)}}\\
				\hline
				PGDs              & 2.870                  & 74.1\%              \\
				~~~ + CW          & 2.617                  & 74.1\%                 \\
				~~~~~~~~~~~~~~~~~ + AutoPGD          & 2.617                  & 74.1\%                   \\
				\hline
				\multicolumn{3} {|c|}{\textbf{(ImageNet) Robust ResNet50 ($\ell_\infty | 4/255$)}}\\
				\hline
				PGDs              & 2.385                  & 98.9\%                  \\
				~~~ + CW          & 2.058                  & 98.9\%                      \\
				~~~~~~~~~~~~~~~~~ + AutoPGD          & 2.058                  & 98.9\%                       \\
				\hline
				\multicolumn{3} {|c|}{\textbf{(ImageNet) Robust ResNet50 ($\ell_\infty | 8/255$)}}\\
				\hline
				PGDs              & 2.378                  & 99.1\%                  \\
				~~~ + CW          & 1.949                  & 99.1\%                  \\
				~~~~~~~~~~~~~~~~~ + AutoPGD          & 1.949                  & 99.1\%                   \\
				\hline
				\multicolumn{3} {|c|}{\textbf{(CIFAR-10) Standard ResNet50}}\\
				\hline
				PGDs              & 0.412                  & 98.7\%                   \\
				~~~ + CW          & 0.120                  & 98.7\%                     \\
				~~~~~~~~~~~~~~~~~ + AutoPGD          & 0.120                  & 98.7\%                      \\
				\hline
				\multicolumn{3} {|c|}{\textbf{(CIFAR-10) Robust ResNet50 ($\ell_2 | 0.5$)}}\\
				\hline
				PGDs              & 1.288                  & 99.9\%                   \\
				~~~ + CW          & 1.096                  & 99.9\%                      \\
				~~~~~~~~~~~~~~~~~ + AutoPGD          & 1.096                  & 99.9\%                   \\
				\hline
			\end{tabular}
			\caption{}
			\label{table:adversary-difference}
		\end{subfigure}
		\hfill
		\begin{subfigure}[b]{0.43\textwidth}
			\centering
			\begin{tabular}{ccccc} \toprule
				\small
					CIFAR10 & \multicolumn{2}{c}{standard} & \multicolumn{2}{c}{robust}  \\ \midrule
					$\epsilon$  &\multicolumn{2}{c}{0.5}& \multicolumn{2}{c}{1.0}\\
					topk & \multicolumn{2}{c}{10}&  \multicolumn{2}{c}{10}\\
					max iters  & \multicolumn{2}{c}{15} & \multicolumn{2}{c}{15}\\ \midrule
					ImageNet & \multicolumn{2}{c}{standard} & \multicolumn{2}{c}{robust}  \\ \midrule
					$\epsilon$  &\multicolumn{2}{c}{2.0}& \multicolumn{2}{c}{6.0}\\
					topk & \multicolumn{2}{c}{15}&  \multicolumn{2}{c}{15}\\
					max iters & \multicolumn{2}{c}{15} & \multicolumn{2}{c}{15}\\ \bottomrule
				\end{tabular}
				\vspace{40pt}
				\caption{}
				\label{table:agi_setup}
		  \end{subfigure}
		   \caption{(a): \textit{Pipeline}: the methods used for boundary search. \textit{Avg Distance}: the average $\ell_2$ distance between the input to the boundary. \textit{Success Rate}: the percentage when the pipeline returns an adversarial example. \textit{Time}: per-instance time with a batch size of 64. We are using much bigger $\epsilon$s for robust models, so the success rates are higher than a standard model. (b): Hyper-parameters used for AGI. We use the default parameteres from the authors' implementation for ImageNet and make minimal changes for CIFAR-10.}
		   \label{fig:attck-performance-and-agi-setup}
	\end{figure}
	
	Our boundary search uses a pipeline of PGDs, CW and AutoPGD. Adversarial examples returned by each method are compared with others and closer ones are returned. If an adversarial example is not found, the pipeline will return the point from the last iteration of the first method (PGDs in our case). Hyper-parameters for each attack can be found in Table~\ref{table:boundary-search-params}. The implementation of PGDs and CW are based on Foolbox~\citep{rauber2017foolboxnative, rauber2017foolbox} and the implementation of AutoPGD is based on the authors' public repository\footnote{\url{https://github.com/fra31/auto-attack}} (we only use \texttt{apgd-ce} and \texttt{apgd-dlr} losses for efficiency reasons). All computations are done using a GPU accelerator Titan RTX with a memory size of 24 GB. Comparisons on the results of the ensemble of these three approaches are shown in Fig.~\ref{table:adversary-difference}.

	\begin{table}[t]
		\centering
		\begin{tabular}{cc|cccc} \toprule
				&CIFAR10 & \multicolumn{2}{c}{standard} & \multicolumn{2}{c}{robust}  \\ \cmidrule{2-6}
				&$\epsilon$s  &\multicolumn{2}{c}{$[0.2, 0.4, 0.6, 0.8, 1.0]$}& \multicolumn{2}{c}{$[0.25, 0.5, 1.0, 1.5, 2.0]$}\\
				&max steps  & \multicolumn{2}{c}{100}&  \multicolumn{2}{c}{100}\\
				&step size  & \multicolumn{2}{c}{5e-3} & \multicolumn{2}{c}{5e-3}\\ \cmidrule{2-6}
				PGDs&ImageNet & \multicolumn{2}{c}{standard} & \multicolumn{2}{c}{robust} \\ \cmidrule{2-6}
				&$\epsilon$s  &\multicolumn{2}{c}{$[36/255., 64/255., 0.3, 0.5, 0.7, 0.9, 1.1]$}& \multicolumn{2}{c}{$[1.0, 2.0, 3.0, 4.0, 5.0, 6.0]$}\\
				&max steps  & \multicolumn{2}{c}{100}&  \multicolumn{2}{c}{100}\\
				&step size  & \multicolumn{2}{c}{\texttt{adaptive}} & \multicolumn{2}{c}{\texttt{adaptive}}\\ \cmidrule{1-6}
				&CIFAR10 & \multicolumn{2}{c}{standard} & \multicolumn{2}{c}{robust}  \\ \cmidrule{2-6}
				&$\epsilon$  &\multicolumn{2}{c}{$1.0$}& \multicolumn{2}{c}{$2.0$}\\
				&max steps  & \multicolumn{2}{c}{100}&  \multicolumn{2}{c}{100}\\
				&step size  & \multicolumn{2}{c}{1e-3} & \multicolumn{2}{c}{1e-3}\\ \cmidrule{2-6}
				CW&ImageNet & \multicolumn{2}{c}{standard} & \multicolumn{2}{c}{robust} \\ \cmidrule{2-6}
				&$\epsilon$  &\multicolumn{2}{c}{$1.0$}& \multicolumn{2}{c}{$6.0$}\\
				&max steps  & \multicolumn{2}{c}{100}&  \multicolumn{2}{c}{100}\\
				&step size  & \multicolumn{2}{c}{1e-2} & \multicolumn{2}{c}{5e-2}\\ \cmidrule{1-6}
				&CIFAR10 & \multicolumn{2}{c}{standard} & \multicolumn{2}{c}{robust}  \\ \cmidrule{2-6}
				&$\epsilon$  &\multicolumn{2}{c}{$1.0$}& \multicolumn{2}{c}{$2.0$}\\
				&max steps  & \multicolumn{2}{c}{100}&  \multicolumn{2}{c}{100}\\
				&step size  & \multicolumn{2}{c}{6e-3} & \multicolumn{2}{c}{1.6e-2}\\ \cmidrule{2-6}
				AutoPGD&ImageNet & \multicolumn{2}{c}{standard} & \multicolumn{2}{c}{robust} \\ \cmidrule{2-6}
				&$\epsilon$  &\multicolumn{2}{c}{$1.1$}& \multicolumn{2}{c}{$6.0$}\\
				&max steps  & \multicolumn{2}{c}{100}&  \multicolumn{2}{c}{100}\\
				&step size  & \multicolumn{2}{c}{2.3e-2} & \multicolumn{2}{c}{1.2e-1}\\\bottomrule
			\end{tabular}
			 \caption{Hyper-parameters used for adversarial attacks. \texttt{adaptive} means the actual step size is determined by $2 * \epsilon$ / max steps.}
			 \label{table:boundary-search-params}
	\end{table}
	
	\subsection{Hyper-parameters for Attribution Methods}\label{appendix:hps-for-attributions}

	All attributions are implemented with Captum~\citep{kokhlikyan2020captum} and visualized with Trulens~\citep{trulens}. For BIG and IG, we use 20 intermediate points between the baseline and the input and the interpolation method is set to \texttt{riemann\_trapezoid}. For AGI, we base on the authors' public repository\footnote{\url{https://github.com/pd90506/AGI}}. The choice of hyper-paramters follow the default choice from the authors for ImageNet and we make minimal changes to adapt them to CIFAR-10 (see Fig.~\ref{table:agi_setup}).
	
	To visualize the attribution map, we use the \texttt{HeatmapVisualizer} with \texttt{blur=10, normalization\_type="signed\_max"} and default values for other keyword arguments from Trulens. 
	
\subsection{Detailed Results on Localization Metrics}\label{sec:appendix-boxplot}

We show the average scores for each localizaiton metrics in Sec.~\ref{sec:evaluation}. We also show the boxplots of the scores for each localization metrics in Fig.~\ref{fig:more-attribution-comparison} for the standard ResNet50 model and Fig.~\ref{fig:robust-more-attribution-comparison} for the robust ResNet50 ($\ell_2|3.0$). All higher scores are better results.

\begin{figure}[!t]
    \centering
    \includegraphics[width=0.7\textwidth]{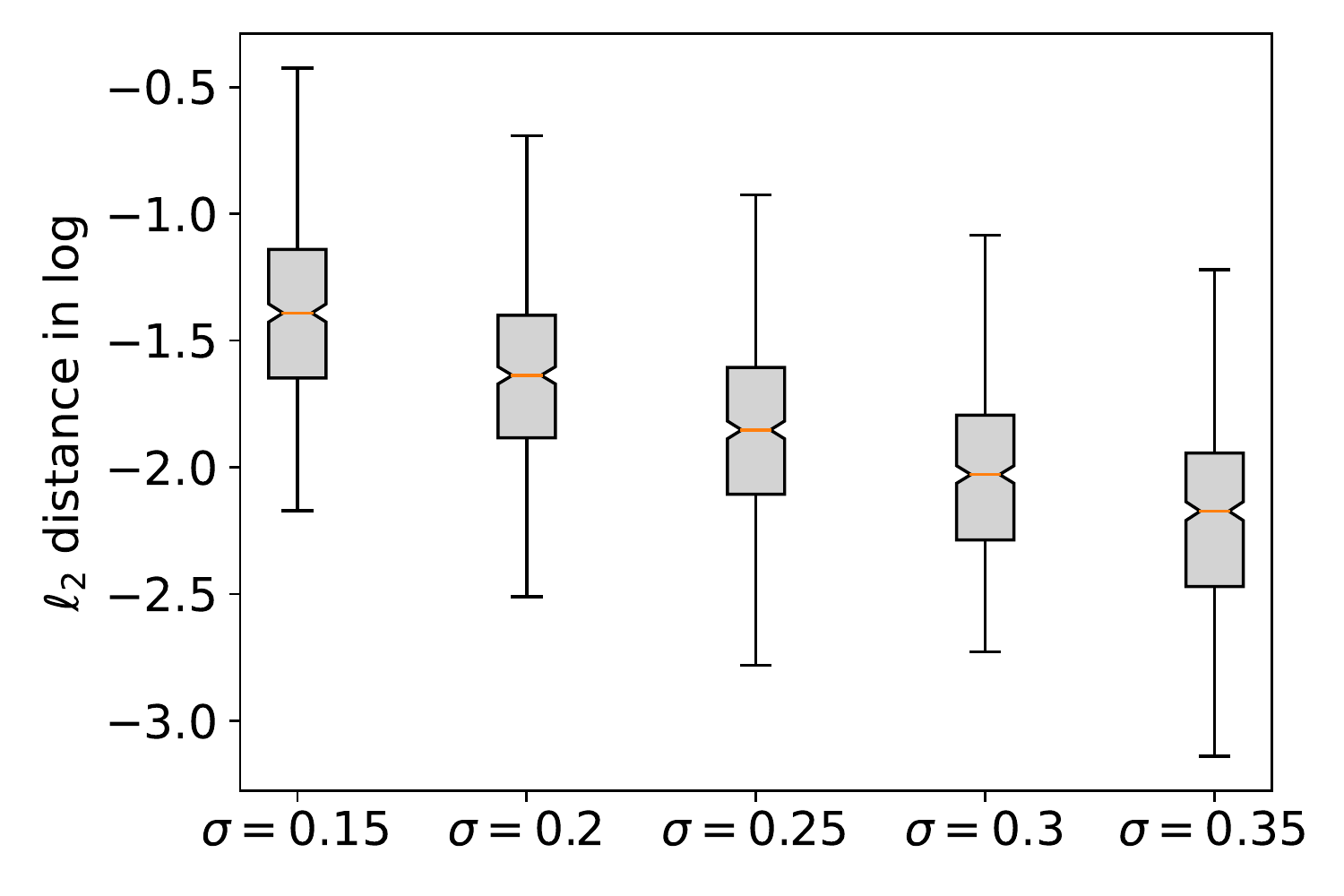}
    \caption{$\ell_2$ distances in logarithm between SG and BSG against different standard deviations $\sigma$ of the Gaussian noise. Results are computed on ResNet50. Notice the first column corresponds to $\sigma=0$.}\label{fig:SG}
	\label{fig:SG}
\end{figure}

\begin{figure}[!t]
	\centering
	\includegraphics[width=0.7\textwidth]{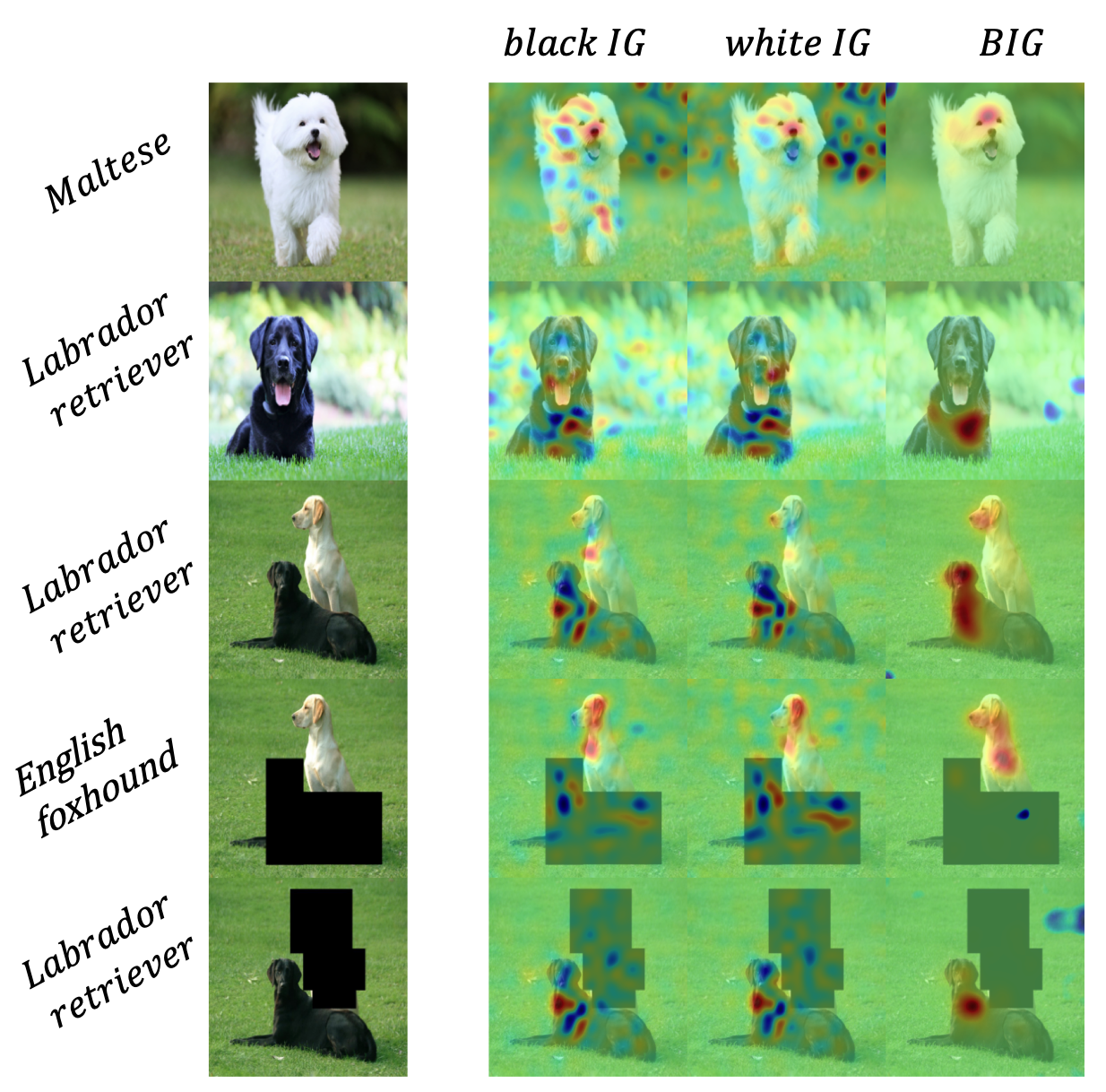}
	\caption{Full results of Fig.~\ref{fig:baseline-demo} in Sec.~\ref{sec:application}. For the third, fourth and fifth example, we compute the attribution scores towards the prediction of the third example, \texttt{{Labrador retriever}.} IG with black or white attributions show that masked area contribute a lot to the prediction while BIG ``accurately'' locate the relevant features in the image with the network's prediction.}\label{fig:baseline_full_comparison}
\end{figure}

\subsection{Additional Experiment with Smoothed Gradient}

We empircally validate Theorem~\ref{theorem-randomized-smoothing} in deeper networks, i.e. ResNet50, which suggests that SM on BSM are more similar on models with randomized smoothing, which are known to be robust~\cite{Cohen2019CertifiedAR}.
To obtain meaningful explanations on smoothed models, which are implemented by evaluating the model on a large set of noised inputs, we assume that the random seed is known by the adversarial example generator, and search for perturbations that point towards a boundary on as many of the noised inputs simultaneously as possible.
We do the boundary search for a subset of 500 images, as this computation is significantly more expensive than previous experiments. 
Instead of directly computing SM and BSM on the smoothed classifier, we utilize the connection between randomized smoothing and SG (see Theorem~\ref{theorem:SG-and-RS} in the Appendix~\ref{appendix:proofs}); therefore, we compare the difference between SG on the clean inputs and SG on their adversarial examples (referred as BSG). The setup is shown as follows.

To generate the adversarial examples for the smoothed classifier of ResNet50 with randomized smoothing, we need to compute back-propagation through the noises. The noise sampler is usually not accessible to the attacker who wants to fool a model with randomized smoothing. However, our goal in this section is not to reproduce the attack with similar setup in practice, instead, what we are after is the point on the boundary. We therefore do the noise sampling prior to run PGD attack, and we use the same noise across all the instances. The steps are listed as follows:
\begin{enumerate}
	\item We use \texttt{numpy.random.randn} as the sampler for Gaussian noise with its random seed set to 2020. We use 50 random noises per instance.
	\item In PGD attack, we aggregate the gradients of all 50 random inputs before we take a regular step to update the input. 
	\item We set $\epsilon=3.0$ and we run at most 40 iterations with a step size of $2 * \epsilon / 40$.
	\item The early stop criteria for the loop of PGD is that when less than 10\% of all randomized points have the original prediction.
	\item When computing Smooth Gradient for the original points or for the adversarial points, we use the same random noise that we generated to approximate the smoothed classifier.
\end{enumerate}

We plot the results in Fig.~\ref{fig:SG}. 
Notably, the trend of the log difference against the standard deviation $\sigma$ used for the Gaussian noise validates that the qualitative meaning of Theorem~\ref{theorem-randomized-smoothing} holds even for large networks.

\section{Symmetry of Attribution Methods}\label{sec:appendix-symmetry}

~\citet{sundararajan2017axiomatic} prove that a linear path is the only path integral that satisifes \emph{symmetry}; that is, when two features' orders are changed for a network that is not using any order information from the input, their attribution scores should not change. One simple way to show the importance of \emph{symmetry} by the following example and we refer \citet{sundararajan2017axiomatic} to readers for more analysis.

\begin{example}
	Consider a function $f(x, y) = min(x, y)$ and to attribute the output of $f$ to the inputs at $x=1, y=1$ we consider a baseline $x=0, y=0$. An example non-linear path from the baseline to the input can be $(x=0, y=0) \rightarrow (x=1, y=0) \rightarrow (x=1, y=1)$. On this path, $f(x, y) = min(x, y) = y $ after the point $(x=1, y=0)$; therefore, gradient integral will return $0$ for the attribution score of $x$ and $1$ for y (we ignore the infinitesimal part of $(x=0, y=0) \rightarrow (x=1, y=0)$). Similarly, when choosing a path $(x=0, y=0) \rightarrow (x=0, y=1) \rightarrow (x=1, y=1)$, we find $x$ is more important. Only the linear path will return 1 for both variables in this case.  
\end{example}

\section{Counterfactual Analysis in the Baseline Selection}\label{appendix-counterfactual-analysis-for-baseline}
	
	The discussion in Sec.~\ref{sec:application} shows an example where there are two dogs in the image. IG with black baseline shows that the body of the white dog is also useful to the model to predict its label and the black dog is a mix: part of the black dog has positive attributions and the rest is negatively contribute to the prediction. However, our proposed method BIG clearly shows that the most important part is the black dog and then comes to the white dog. To validate where the model is actually using the white dog, we manually remove the black dog or the white dog from the image and see if the model retain its prediction. The result is shown in Fig.~\ref{fig:baseline_full_comparison}. Clearly, when removing the black dog, the model changes its prediction from \texttt{Labrador retriever} to \texttt{English foxhound} while removing the white dog does not change the prediction. This result helps to convince the reader that BIG is more reliable than IG with black baseline in this case as a more faithful explanation to the classification result for this instance. 

\section{Additional Visualizations for BIG}\label{appendix-visualizations-for-big}

\begin{figure}[!t]
	\centering
	\includegraphics[width=1\textwidth]{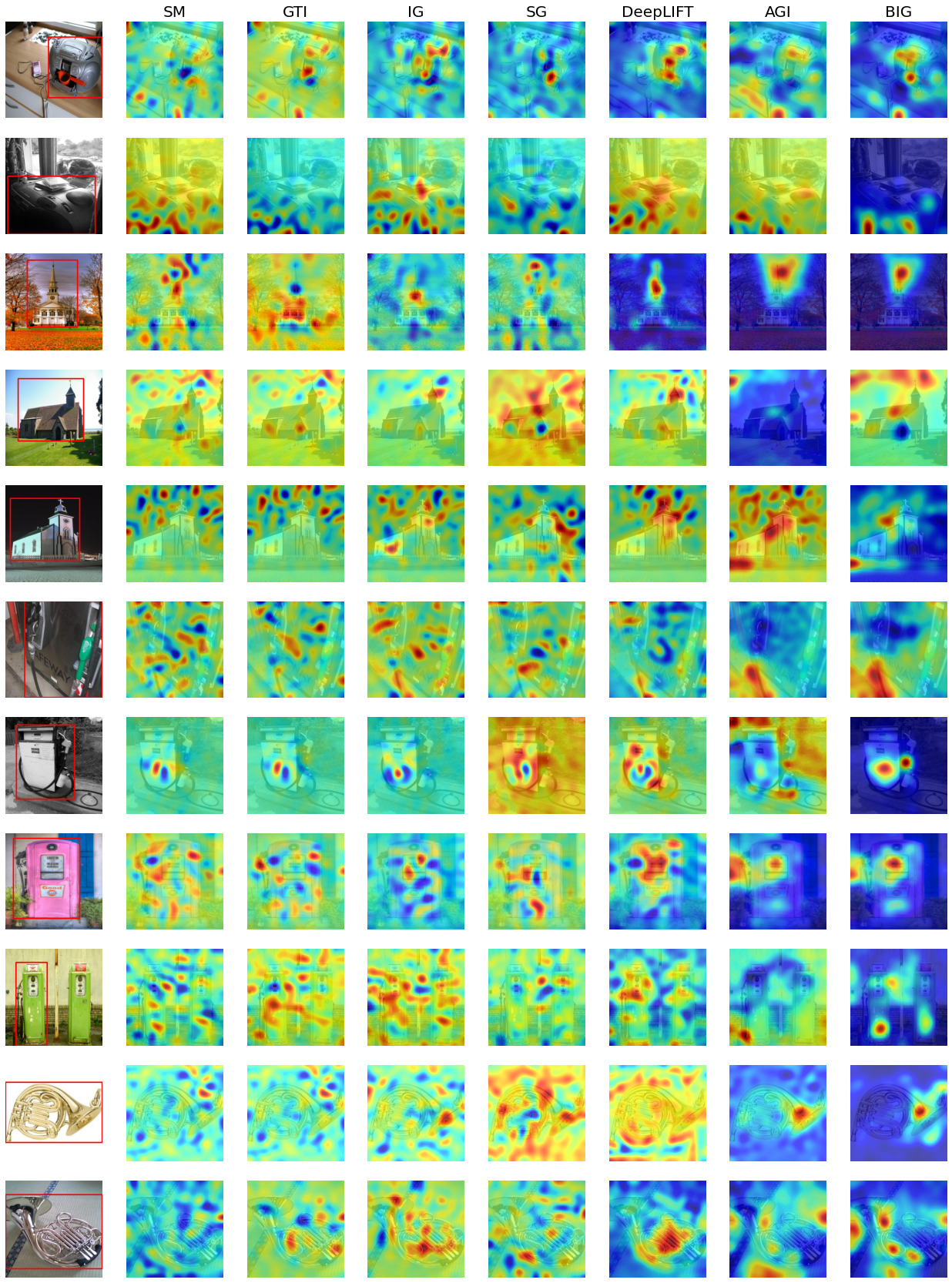}
	\caption{Visualizations of different attributions for a standard ResNet50}
	\label{fig:more-visualizations}
\end{figure}

\begin{figure}[!t]
	\centering
	\includegraphics[width=1\textwidth]{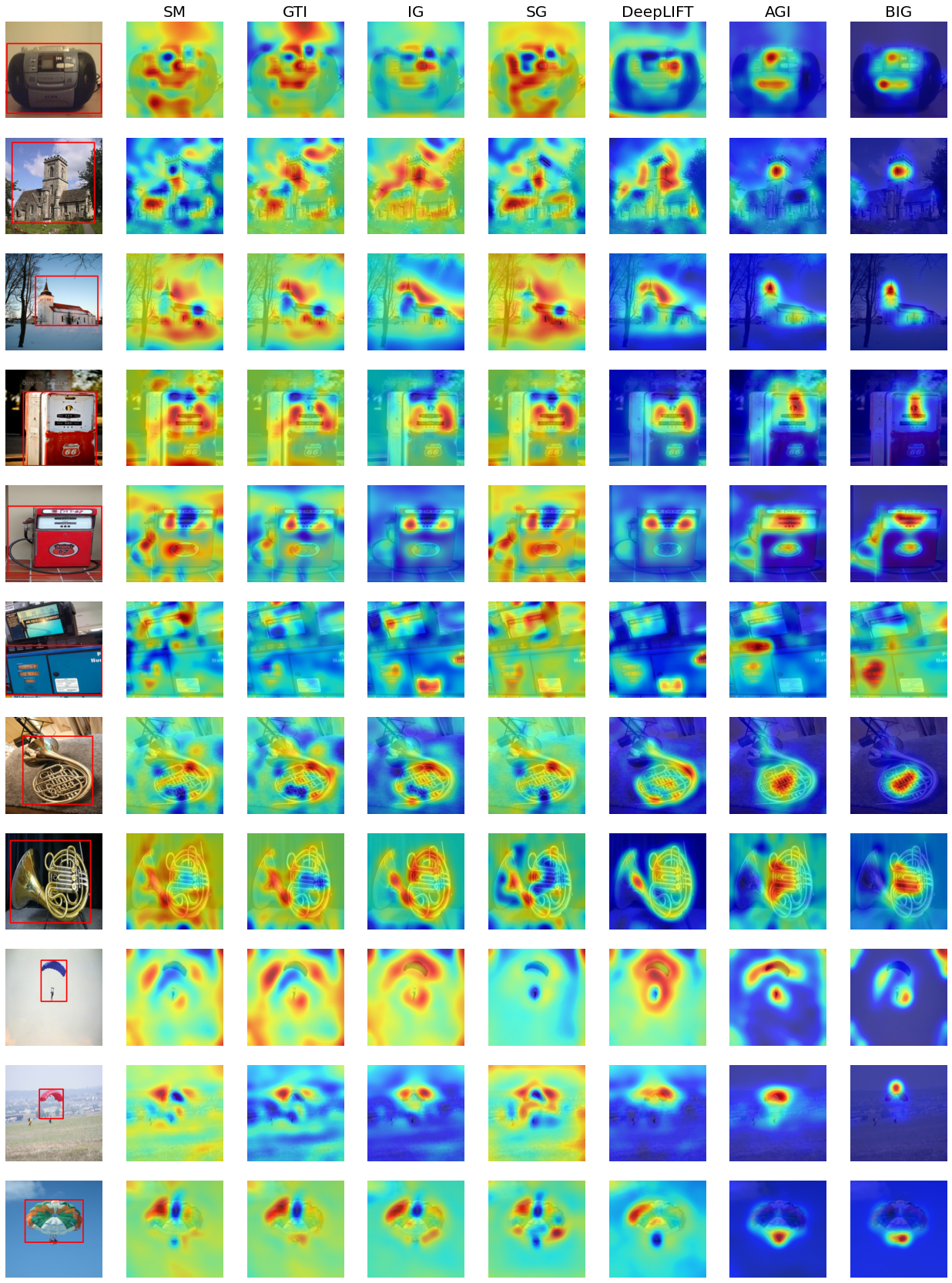}
	\caption{Visualizations of different attributions for a robust ($\ell_2|3.0$) ResNet50}
	\label{fig:more-robust-visualizations}
\end{figure}

More visualizations comparing BIG with other attributions can be found in Fig.~\ref{fig:more-visualizations} and ~\ref{fig:more-robust-visualizations}.

%% file: main.bbl
\begin{thebibliography}{59}
\providecommand{\natexlab}[1]{#1}
\providecommand{\url}[1]{\texttt{#1}}
\expandafter\ifx\csname urlstyle\endcsname\relax
  \providecommand{\doi}[1]{doi: #1}\else
  \providecommand{\doi}{doi: \begingroup \urlstyle{rm}\Url}\fi

\bibitem[Adebayo et~al.(2018)Adebayo, Gilmer, Muelly, Goodfellow, Hardt, and
  Kim]{adebayo2018sanity}
Julius Adebayo, Justin Gilmer, Michael Muelly, Ian Goodfellow, Moritz Hardt,
  and Been Kim.
\newblock Sanity checks for saliency maps.
\newblock In \emph{Advances in Neural Information Processing Systems}, 2018.

\bibitem[Adebayo et~al.(2020)Adebayo, Muelly, Liccardi, and
  Kim]{adebayo2020debugging}
Julius Adebayo, Michael Muelly, Ilaria Liccardi, and Been Kim.
\newblock Debugging tests for model explanations, 2020.

\bibitem[Ancona et~al.(2017)Ancona, Ceolini, Öztireli, and
  Gross]{ancona2017better}
Marco Ancona, Enea Ceolini, Cengiz Öztireli, and Markus Gross.
\newblock Towards better understanding of gradient-based attribution methods
  for deep neural networks, 2017.

\bibitem[Ancona et~al.(2018)Ancona, Ceolini, Öztireli, and
  Gross]{ancona2018towards}
Marco Ancona, Enea Ceolini, Cengiz Öztireli, and Markus Gross.
\newblock Towards better understanding of gradient-based attribution methods
  for deep neural networks.
\newblock In \emph{International Conference on Learning Representations}, 2018.

\bibitem[Binder et~al.(2016)Binder, Montavon, Lapuschkin, M{\"u}ller, and
  Samek]{binder2016layer}
Alexander Binder, Gr{\'e}goire Montavon, Sebastian Lapuschkin, Klaus-Robert
  M{\"u}ller, and Wojciech Samek.
\newblock Layer-wise relevance propagation for neural networks with local
  renormalization layers.
\newblock In \emph{International Conference on Artificial Neural Networks},
  pp.\  63--71. Springer, 2016.

\bibitem[Carlini \& Wagner(2017)Carlini and Wagner]{Carlini2017TowardsET}
Nicholas Carlini and D.~Wagner.
\newblock Towards evaluating the robustness of neural networks.
\newblock \emph{2017 IEEE Symposium on Security and Privacy (SP)}, pp.\
  39--57, 2017.

\bibitem[Chalasani et~al.(2020)Chalasani, Chen, Chowdhury, Wu, and
  Jha]{chalasani2020concise}
Prasad Chalasani, Jiefeng Chen, Amrita~Roy Chowdhury, Xi~Wu, and Somesh Jha.
\newblock Concise explanations of neural networks using adversarial training.
\newblock In \emph{International Conference on Machine Learning}, pp.\
  1383--1391. PMLR, 2020.

\bibitem[Chang et~al.(2019)Chang, Creager, Goldenberg, and
  Duvenaud]{Chang2019ExplainingIC}
Chun-Hao Chang, Elliot Creager, Anna Goldenberg, and D.~Duvenaud.
\newblock Explaining image classifiers by counterfactual generation.
\newblock In \emph{ICLR}, 2019.

\bibitem[Chattopadhyay et~al.(2017)Chattopadhyay, Sarkar, Howlader, and
  Balasubramanian]{chattopadhyay2017grad}
Aditya Chattopadhyay, Anirban Sarkar, Prantik Howlader, and Vineeth~N
  Balasubramanian.
\newblock Grad-cam++: Generalized gradient-based visual explanations for deep
  convolutional networks.
\newblock \emph{arXiv preprint arXiv:1710.11063}, 2017.

\bibitem[Chen et~al.(2019)Chen, Wu, Rastogi, Liang, and
  Jha]{NEURIPS2019_172ef5a9}
Jiefeng Chen, Xi~Wu, Vaibhav Rastogi, Yingyu Liang, and Somesh Jha.
\newblock Robust attribution regularization.
\newblock In \emph{Advances in Neural Information Processing Systems}, 2019.

\bibitem[Cohen et~al.(2019)Cohen, Rosenfeld, and Kolter]{Cohen2019CertifiedAR}
Jeremy~M. Cohen, Elan Rosenfeld, and J.~Z. Kolter.
\newblock Certified adversarial robustness via randomized smoothing.
\newblock In \emph{ICML}, 2019.

\bibitem[Croce \& Hein(2020)Croce and Hein]{croce2020reliable}
Francesco Croce and Matthias Hein.
\newblock Reliable evaluation of adversarial robustness with an ensemble of
  diverse parameter-free attacks.
\newblock In \emph{ICML}, 2020.

\bibitem[Croce et~al.(2019)Croce, Andriushchenko, and Hein]{croce2019provable}
Francesco Croce, Maksym Andriushchenko, and Matthias Hein.
\newblock Provable robustness of relu networks via maximization of linear
  regions.
\newblock \emph{AISTATS 2019}, 2019.

\bibitem[Dhamdhere et~al.(2019)Dhamdhere, Sundararajan, and
  Yan]{dhamdhere2018how}
Kedar Dhamdhere, Mukund Sundararajan, and Qiqi Yan.
\newblock How important is a neuron.
\newblock In \emph{International Conference on Learning Representations}, 2019.
\newblock URL \url{https://openreview.net/forum?id=SylKoo0cKm}.

\bibitem[Dhurandhar et~al.(2018)Dhurandhar, Chen, Luss, Tu, Ting, Shanmugam,
  and Das]{Dhurandhar2018ExplanationsBO}
A.~Dhurandhar, P.~Chen, Ronny Luss, Chun-Chen Tu, Pai-Shun Ting, Karthikeyan
  Shanmugam, and Payel Das.
\newblock Explanations based on the missing: Towards contrastive explanations
  with pertinent negatives.
\newblock In \emph{NeurIPS}, 2018.

\bibitem[Dombrowski et~al.(2019)Dombrowski, Alber, Anders, Ackermann,
  M{\"u}ller, and Kessel]{Dombrowski2019ExplanationsCB}
Ann-Kathrin Dombrowski, M.~Alber, Christopher~J. Anders, Marcel Ackermann,
  K.~M{\"u}ller, and P.~Kessel.
\newblock Explanations can be manipulated and geometry is to blame.
\newblock In \emph{NeurIPS}, 2019.

\bibitem[Engstrom et~al.(2019)Engstrom, Ilyas, Salman, Santurkar, and
  Tsipras]{robustness}
Logan Engstrom, Andrew Ilyas, Hadi Salman, Shibani Santurkar, and Dimitris
  Tsipras.
\newblock Robustness (python library), 2019.
\newblock URL \url{https://github.com/MadryLab/robustness}.

\bibitem[Etmann et~al.(2019)Etmann, Lunz, Maass, and
  Schoenlieb]{pmlr-v97-etmann19a}
Christian Etmann, Sebastian Lunz, Peter Maass, and Carola Schoenlieb.
\newblock On the connection between adversarial robustness and saliency map
  interpretability.
\newblock In \emph{Proceedings of the 36th International Conference on Machine
  Learning}, 2019.

\bibitem[{Fong} \& {Vedaldi}(2017){Fong} and {Vedaldi}]{8237633}
R.~C. {Fong} and A.~{Vedaldi}.
\newblock Interpretable explanations of black boxes by meaningful perturbation.
\newblock In \emph{2017 IEEE International Conference on Computer Vision
  (ICCV)}, pp.\  3449--3457, 2017.
\newblock \doi{10.1109/ICCV.2017.371}.

\bibitem[Fong \& Vedaldi(2017)Fong and Vedaldi]{fong2017interpretable}
Ruth~C Fong and Andrea Vedaldi.
\newblock Interpretable explanations of black boxes by meaningful perturbation.
\newblock In \emph{Proceedings of the IEEE international conference on computer
  vision}, pp.\  3429--3437, 2017.

\bibitem[Fromherz et~al.(2021)Fromherz, Leino, Fredrikson, Parno, and
  Păsăreanu]{fromherz20projections}
Aymeric Fromherz, Klas Leino, Matt Fredrikson, Bryan Parno, and Corina
  Păsăreanu.
\newblock Fast geometric projections for local robustness certification.
\newblock In \emph{International Conference on Learning Representations
  (ICLR)}, 2021.

\bibitem[Ghalebikesabi et~al.(2021)Ghalebikesabi, Ter-Minassian, Diaz-Ordaz,
  and Holmes]{Ghalebikesabi2021OnLO}
Sahra Ghalebikesabi, Lucile Ter-Minassian, Karla Diaz-Ordaz, and Chris~C.
  Holmes.
\newblock On locality of local explanation models.
\newblock \emph{arxiv}, abs/2106.14648, 2021.

\bibitem[Goodfellow et~al.(2015)Goodfellow, Shlens, and Szegedy]{43405}
Ian Goodfellow, Jonathon Shlens, and Christian Szegedy.
\newblock Explaining and harnessing adversarial examples.
\newblock In \emph{ICLR}, 2015.

\bibitem[Goyal et~al.(2019)Goyal, Wu, Ernst, Batra, Parikh, and
  Lee]{pmlr-v97-goyal19a}
Yash Goyal, Ziyan Wu, Jan Ernst, Dhruv Batra, Devi Parikh, and Stefan Lee.
\newblock Counterfactual visual explanations.
\newblock In \emph{Proceedings of the 36th International Conference on Machine
  Learning}, pp.\  2376--2384, 2019.

\bibitem[Howard()]{imagenette}
Jeremy Howard.
\newblock imagenette.
\newblock URL \url{https://github.com/fastai/imagenette/}.

\bibitem[Ilyas et~al.(2019)Ilyas, Santurkar, Tsipras, Engstrom, Tran, and
  Madry]{NEURIPS2019_e2c420d9}
Andrew Ilyas, Shibani Santurkar, Dimitris Tsipras, Logan Engstrom, Brandon
  Tran, and Aleksander Madry.
\newblock Adversarial examples are not bugs, they are features.
\newblock In \emph{Advances in Neural Information Processing Systems}, 2019.

\bibitem[Jordan et~al.(2019)Jordan, Lewis, and Dimakis]{Jordan2019ProvableCF}
Matt Jordan, J.~Lewis, and A.~Dimakis.
\newblock Provable certificates for adversarial examples: Fitting a ball in the
  union of polytopes.
\newblock In \emph{NeurIPS}, 2019.

\bibitem[Kim et~al.(2018)Kim, Wattenberg, Gilmer, Cai, Wexler, Vi{\'e}gas, and
  Sayres]{Kim2018InterpretabilityBF}
Been Kim, M.~Wattenberg, J.~Gilmer, C.~J. Cai, James Wexler, F.~Vi{\'e}gas, and
  Rory Sayres.
\newblock Interpretability beyond feature attribution: Quantitative testing
  with concept activation vectors (tcav).
\newblock In \emph{ICML}, 2018.

\bibitem[Kokhlikyan et~al.(2020)Kokhlikyan, Miglani, Martin, Wang, Alsallakh,
  Reynolds, Melnikov, Kliushkina, Araya, Yan, and
  Reblitz-Richardson]{kokhlikyan2020captum}
Narine Kokhlikyan, Vivek Miglani, Miguel Martin, Edward Wang, Bilal Alsallakh,
  Jonathan Reynolds, Alexander Melnikov, Natalia Kliushkina, Carlos Araya, Siqi
  Yan, and Orion Reblitz-Richardson.
\newblock Captum: A unified and generic model interpretability library for
  pytorch, 2020.

\bibitem[Kolter \& Wong(2018)Kolter and Wong]{Kolter2018ProvableDA}
J.~Z. Kolter and E.~Wong.
\newblock Provable defenses against adversarial examples via the convex outer
  adversarial polytope.
\newblock In \emph{ICML}, 2018.

\bibitem[Krizhevsky et~al.()Krizhevsky, Nair, and Hinton]{cifar}
Alex Krizhevsky, Vinod Nair, and Geoffrey Hinton.
\newblock Cifar-10 (canadian institute for advanced research).
\newblock URL \url{http://www.cs.toronto.edu/~kriz/cifar.html}.

\bibitem[Lee et~al.(2020)Lee, Lee, and Park]{lee2020lipschitz}
Sungyoon Lee, Jaewook Lee, and Saerom Park.
\newblock Lipschitz-certifiable training with a tight outer bound.
\newblock \emph{Advances in Neural Information Processing Systems}, 33, 2020.

\bibitem[Leino et~al.(2018)Leino, Sen, Datta, Fredrikson, and
  Li]{leino2018influence}
Klas Leino, Shayak Sen, Anupam Datta, Matt Fredrikson, and Linyi Li.
\newblock Influence-directed explanations for deep convolutional networks.
\newblock In \emph{2018 IEEE International Test Conference (ITC)}, pp.\  1--8.
  IEEE, 2018.

\bibitem[Leino et~al.(2021{\natexlab{a}})Leino, Shih, Fredrikson, She, Wang,
  Lu, Sen, Gopinath, and {, Anupam}]{trulens}
Klas Leino, Ricardo Shih, Matt Fredrikson, Jennifer She, Zifan Wang, Caleb Lu,
  Shayak Sen, Divya Gopinath, and {, Anupam}.
\newblock truera/trulens: Trulens, 2021{\natexlab{a}}.
\newblock URL \url{https://zenodo.org/record/4495856}.

\bibitem[Leino et~al.(2021{\natexlab{b}})Leino, Wang, and
  Fredrikson]{leino2021globallyrobust}
Klas Leino, Zifan Wang, and Matt Fredrikson.
\newblock Globally-robust neural networks, 2021{\natexlab{b}}.

\bibitem[Lundberg \& Lee(2017)Lundberg and Lee]{lundberg2017unified}
Scott~M Lundberg and Su-In Lee.
\newblock A unified approach to interpreting model predictions.
\newblock In \emph{Proceedings of the 31st international conference on neural
  information processing systems}, pp.\  4768--4777, 2017.

\bibitem[Madry et~al.(2018)Madry, Makelov, Schmidt, Tsipras, and
  Vladu]{madry2018towards}
Aleksander Madry, Aleksandar Makelov, Ludwig Schmidt, Dimitris Tsipras, and
  Adrian Vladu.
\newblock Towards deep learning models resistant to adversarial attacks.
\newblock In \emph{International Conference on Learning Representations}, 2018.

\bibitem[Montavon et~al.(2015)Montavon, Bach, Binder, Samek, and
  Müller]{montavon2015explaining}
Grégoire Montavon, Sebastian Bach, Alexander Binder, Wojciech Samek, and
  Klaus-Robert Müller.
\newblock Explaining nonlinear classification decisions with deep taylor
  decomposition, 2015.

\bibitem[Olah et~al.(2017)Olah, Mordvintsev, and Schubert]{olah2017feature}
Chris Olah, Alexander Mordvintsev, and Ludwig Schubert.
\newblock Feature visualization.
\newblock \emph{Distill}, 2017.
\newblock \doi{10.23915/distill.00007}.
\newblock https://distill.pub/2017/feature-visualization.

\bibitem[Pan et~al.(2021)Pan, Li, and Zhu]{pan2021explaining}
Deng Pan, Xin Li, and Dongxiao Zhu.
\newblock Explaining deep neural network models with adversarial gradient
  integration.
\newblock In \emph{Thirtieth International Joint Conference on Artificial
  Intelligence (IJCAI)}, 2021.

\bibitem[Rauber et~al.(2017)Rauber, Brendel, and Bethge]{rauber2017foolbox}
Jonas Rauber, Wieland Brendel, and Matthias Bethge.
\newblock Foolbox: A python toolbox to benchmark the robustness of machine
  learning models.
\newblock In \emph{Reliable Machine Learning in the Wild Workshop, 34th
  International Conference on Machine Learning}, 2017.
\newblock URL \url{http://arxiv.org/abs/1707.04131}.

\bibitem[Rauber et~al.(2020)Rauber, Zimmermann, Bethge, and
  Brendel]{rauber2017foolboxnative}
Jonas Rauber, Roland Zimmermann, Matthias Bethge, and Wieland Brendel.
\newblock Foolbox native: Fast adversarial attacks to benchmark the robustness
  of machine learning models in pytorch, tensorflow, and jax.
\newblock \emph{Journal of Open Source Software}, 5\penalty0 (53):\penalty0
  2607, 2020.
\newblock \doi{10.21105/joss.02607}.
\newblock URL \url{https://doi.org/10.21105/joss.02607}.

\bibitem[Russakovsky et~al.(2015)Russakovsky, Deng, Su, Krause, Satheesh, Ma,
  Huang, Karpathy, Khosla, Bernstein, Berg, and Fei-Fei]{ILSVRC15}
Olga Russakovsky, Jia Deng, Hao Su, Jonathan Krause, Sanjeev Satheesh, Sean Ma,
  Zhiheng Huang, Andrej Karpathy, Aditya Khosla, Michael Bernstein,
  Alexander~C. Berg, and Li~Fei-Fei.
\newblock {ImageNet Large Scale Visual Recognition Challenge}.
\newblock \emph{International Journal of Computer Vision (IJCV)}, 115\penalty0
  (3):\penalty0 211--252, 2015.
\newblock \doi{10.1007/s11263-015-0816-y}.

\bibitem[Samek et~al.(2016)Samek, Binder, Montavon, Lapuschkin, and
  M{\"u}ller]{samek2016evaluating}
Wojciech Samek, Alexander Binder, Gr{\'e}goire Montavon, Sebastian Lapuschkin,
  and Klaus-Robert M{\"u}ller.
\newblock Evaluating the visualization of what a deep neural network has
  learned.
\newblock \emph{IEEE transactions on neural networks and learning systems},
  28\penalty0 (11):\penalty0 2660--2673, 2016.

\bibitem[Selvaraju et~al.(2017)Selvaraju, Cogswell, Das, Vedantam, Parikh, and
  Batra]{selvaraju2017grad}
Ramprasaath~R Selvaraju, Michael Cogswell, Abhishek Das, Ramakrishna Vedantam,
  Devi Parikh, and Dhruv Batra.
\newblock Grad-cam: Visual explanations from deep networks via gradient-based
  localization.
\newblock In \emph{Proceedings of the IEEE international conference on computer
  vision}, pp.\  618--626, 2017.

\bibitem[Shrikumar et~al.(2017)Shrikumar, Greenside, and
  Kundaje]{shrikumar2017learning}
Avanti Shrikumar, Peyton Greenside, and Anshul Kundaje.
\newblock Learning important features through propagating activation
  differences.
\newblock In \emph{International Conference on Machine Learning}, pp.\
  3145--3153. PMLR, 2017.

\bibitem[Simonyan et~al.(2013)Simonyan, Vedaldi, and
  Zisserman]{simonyan2013deep}
Karen Simonyan, Andrea Vedaldi, and Andrew Zisserman.
\newblock Deep inside convolutional networks: Visualising image classification
  models and saliency maps, 2013.

\bibitem[Sinha et~al.(2020)Sinha, Namkoong, Volpi, and
  Duchi]{sinha2020certifying}
Aman Sinha, Hongseok Namkoong, Riccardo Volpi, and John Duchi.
\newblock Certifying some distributional robustness with principled adversarial
  training, 2020.

\bibitem[Smilkov et~al.(2017)Smilkov, Thorat, Kim, Viégas, and
  Wattenberg]{smilkov2017smoothgrad}
Daniel Smilkov, Nikhil Thorat, Been Kim, Fernanda Viégas, and Martin
  Wattenberg.
\newblock Smoothgrad: removing noise by adding noise, 2017.

\bibitem[Springenberg et~al.(2014)Springenberg, Dosovitskiy, Brox, and
  Riedmiller]{springenberg2014striving}
Jost~Tobias Springenberg, Alexey Dosovitskiy, Thomas Brox, and Martin
  Riedmiller.
\newblock Striving for simplicity: The all convolutional net, 2014.

\bibitem[Sturmfels et~al.(2020)Sturmfels, Lundberg, and
  Lee]{sturmfels2020visualizing}
Pascal Sturmfels, Scott Lundberg, and Su-In Lee.
\newblock Visualizing the impact of feature attribution baselines.
\newblock \emph{Distill}, 2020.
\newblock \doi{10.23915/distill.00022}.
\newblock https://distill.pub/2020/attribution-baselines.

\bibitem[Sundararajan et~al.(2017)Sundararajan, Taly, and
  Yan]{sundararajan2017axiomatic}
Mukund Sundararajan, Ankur Taly, and Qiqi Yan.
\newblock Axiomatic attribution for deep networks.
\newblock In \emph{Proceedings of the 34th International Conference on Machine
  Learning-Volume 70}, pp.\  3319--3328. JMLR. org, 2017.

\bibitem[Tjeng et~al.(2019)Tjeng, Xiao, and Tedrake]{tjeng2018evaluating}
Vincent Tjeng, Kai~Y. Xiao, and Russ Tedrake.
\newblock Evaluating robustness of neural networks with mixed integer
  programming.
\newblock In \emph{International Conference on Learning Representations}, 2019.

\bibitem[Wang et~al.(2020{\natexlab{a}})Wang, Wang, Du, Yang, Zhang, Ding,
  Mardziel, and Hu]{wang2020score}
Haofan Wang, Zifan Wang, Mengnan Du, Fan Yang, Zijian Zhang, Sirui Ding, Piotr
  Mardziel, and Xia Hu.
\newblock Score-cam: Score-weighted visual explanations for convolutional
  neural networks.
\newblock In \emph{Proceedings of the IEEE/CVF Conference on Computer Vision
  and Pattern Recognition Workshops}, pp.\  24--25, 2020{\natexlab{a}}.

\bibitem[Wang et~al.(2020{\natexlab{b}})Wang, Mardziel, Datta, and
  Fredrikson]{wang2020interpreting}
Zifan Wang, Piotr Mardziel, Anupam Datta, and Matt Fredrikson.
\newblock Interpreting interpretations: Organizing attribution methods by
  criteria.
\newblock In \emph{Proceedings of the IEEE/CVF Conference on Computer Vision
  and Pattern Recognition Workshops}, pp.\  10--11, 2020{\natexlab{b}}.

\bibitem[Wang et~al.(2020{\natexlab{c}})Wang, Wang, Ramkumar, Mardziel,
  Fredrikson, and Datta]{NEURIPS2020_9d94c898}
Zifan Wang, Haofan Wang, Shakul Ramkumar, Piotr Mardziel, Matt Fredrikson, and
  Anupam Datta.
\newblock Smoothed geometry for robust attribution.
\newblock In H.~Larochelle, M.~Ranzato, R.~Hadsell, M.~F. Balcan, and H.~Lin
  (eds.), \emph{Advances in Neural Information Processing Systems}, volume~33,
  pp.\  13623--13634. Curran Associates, Inc., 2020{\natexlab{c}}.

\bibitem[Weng et~al.(2018)Weng, Zhang, Chen, Song, Hsieh, Boning, Dhillon, and
  Daniel]{Weng2018TowardsFC}
Tsui-Wei Weng, Huan Zhang, H.~Chen, Zhao Song, C.~Hsieh, D.~Boning, I.~Dhillon,
  and L.~Daniel.
\newblock Towards fast computation of certified robustness for relu networks.
\newblock In \emph{ICML}, 2018.

\bibitem[Yang et~al.(2020)Yang, Duan, Hu, Salman, Razenshteyn, and
  Li]{Yang2020RandomizedSO}
Greg Yang, T.~Duan, Edward~J. Hu, Hadi Salman, Ilya~P. Razenshteyn, and
  Jungshian Li.
\newblock Randomized smoothing of all shapes and sizes.
\newblock \emph{ArXiv}, abs/2002.08118, 2020.

\bibitem[Yeh et~al.(2019)Yeh, Hsieh, Suggala, Inouye, and
  Ravikumar]{yeh2019fidelity}
Chih-Kuan Yeh, Cheng-Yu Hsieh, Arun~Sai Suggala, David~I Inouye, and Pradeep
  Ravikumar.
\newblock On the (in) fidelity and sensitivity for explanations.
\newblock In \emph{Advances in Neural Information Processing Systems}, 2019.

\end{thebibliography}
